\documentclass[10pt,letterpaper]{article}

\usepackage[pagenumbers]{cvpr} 
\usepackage[accsupp]{axessibility}  
\usepackage{graphicx}
\usepackage{amsmath}
\usepackage{amssymb}
\usepackage{lipsum}
\usepackage{caption}
\usepackage{booktabs}
\usepackage{xspace}
\usepackage{multirow}
\usepackage[dvipsnames]{xcolor}
\usepackage{float}
\usepackage[numbers]{natbib}
\usepackage{makecell}
\usepackage{flushend,cuted}

\usepackage[pagebackref,breaklinks,colorlinks]{hyperref}

\RequirePackage{enumitem}
\setlist[itemize]{nosep}

\renewcommand{\paragraph}[1]{\vspace{0.2em}\noindent \textbf{#1 \hspace{0.2em}}}

\usepackage[capitalize]{cleveref}
\crefname{section}{Sec.}{Secs.}
\Crefname{section}{Section}{Sections}
\Crefname{table}{Table}{Tables}

\RequirePackage{enumitem}
\setlist[itemize]{nosep}

\usepackage{xcolor}
\newcounter{todos}
\AtEndDocument{\ifnum\value{todos}>0 \PackageWarning{TODOS}{There are \arabic{todos} todos left in this paper! Fix them before submitting the paper!} \fi}

\definecolor{blue-violet}{rgb}{0.54, 0.17, 0.89}

\newcommand{\Netname}{SeSDF\xspace}

\def\cvprPaperID{5448} 
\def\confName{CVPR}
\def\confYear{2023}

\begin{document}
\begin{sloppypar}
\twocolumn

\title{SeSDF: Self-evolved Signed Distance Field for Implicit 3D Clothed Human Reconstruction}

\author{Yukang Cao
\quad\quad
Kai Han
\quad\quad
Kwan-Yee K. Wong
\\
The University of Hong Kong\\
{\tt\small \{ykcao, kykwong\}@cs.hku.hk \qquad kaihanx@hku.hk}
}
\twocolumn[{
\renewcommand\twocolumn[1][]{#1}
\maketitle
\begin{center}
    \captionsetup{type=figure}
    \includegraphics[width=0.85\linewidth]{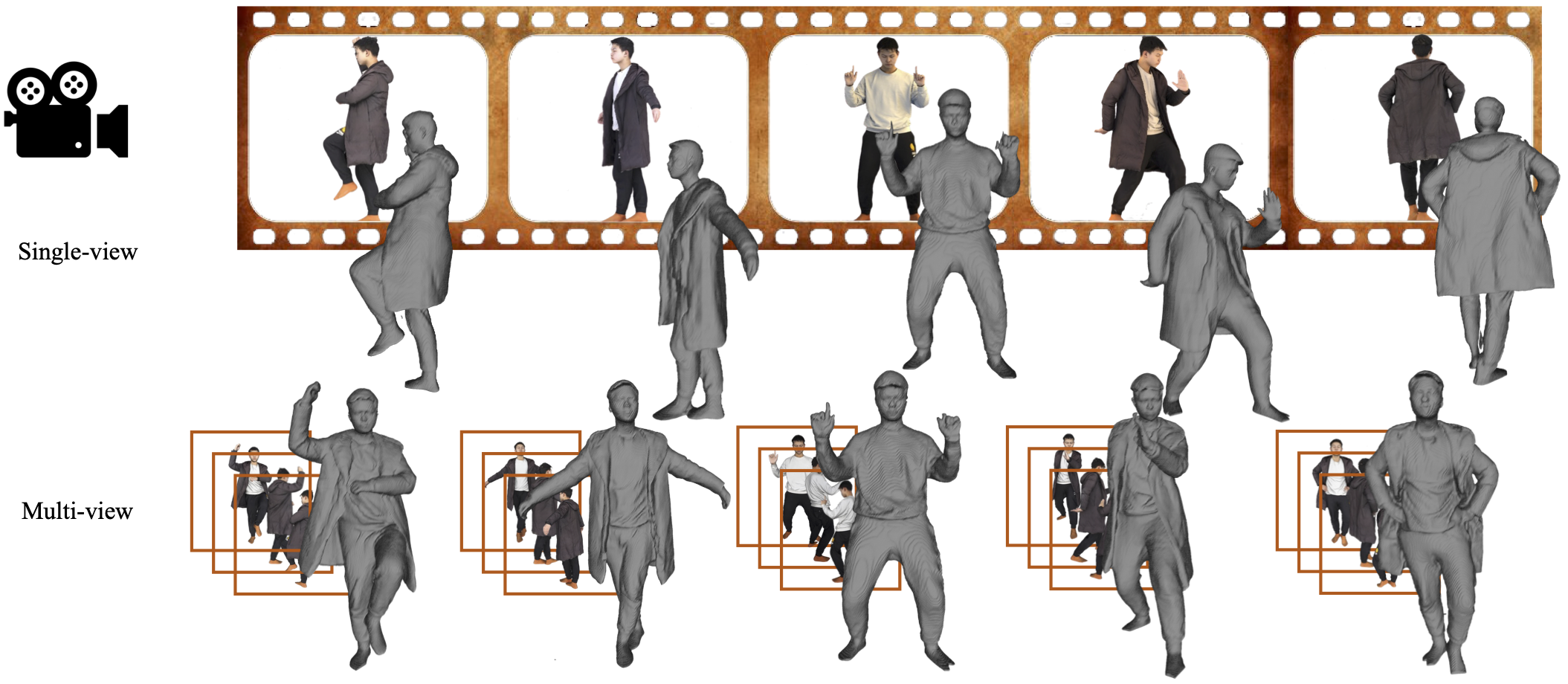}
    \captionof{figure}{\textbf{Self-evolved Signed Distance Field (SeSDF): }We propose Self-evolved Signed Distance Field (SeSDF), which can flexibly reconstruct 3D clothed human models from a single-view image (top) or uncalibrated multi-view images (bottom). \Netname can robustly recover fine geometry details from any type of poses, which allows us to generate clothed human avatars.}
  \label{fig:teaser}
\end{center}
}]

\begin{abstract}
We address the problem of clothed human reconstruction from a single image or uncalibrated multi-view images. Existing methods struggle with reconstructing detailed geometry of a clothed human and often require a calibrated setting for multi-view reconstruction. We propose a flexible framework which, by leveraging the parametric SMPL-X model, can take an arbitrary number of input images to reconstruct a clothed human model under an uncalibrated setting. At the core of our framework is our novel self-evolved signed distance field (\Netname) module which allows the framework to learn to deform the signed distance field (SDF) derived from the fitted SMPL-X model, such that detailed geometry reflecting the actual clothed human can be encoded for better reconstruction. Besides, we propose a simple method for self-calibration of multi-view images via the fitted SMPL-X parameters. This lifts the requirement of tedious manual calibration and largely increases the flexibility of our method. Further, we introduce an effective occlusion-aware feature fusion strategy to account for the most useful features to reconstruct the human model. We thoroughly evaluate our framework on public benchmarks, demonstrating significant superiority over the state-of-the-arts both qualitatively and quantitatively. 

\end{abstract}

\vspace{-1em}
\section{Introduction}
\label{sec:intro}
Clothed human reconstruction is a hot topic with increasing demand in real-world applications such as 3D telepresence, game modeling, metaverse~\cite{mystakidis2022metaverse}, etc.
Early works show promise under equipment-assisted settings, requiring expensive dense camera rigs~\cite{Collet2015HighqualitySF} and tedious calibration procedures~\cite{favalli2012multiview}. 
Parametric models, such as SMPL~\cite{SMPL:2015} and SMPL-X~\cite{SMPL-X:2019}, have been introduced to model a naked human body with constrained parameters. With the witnessed success of deep learning in many vision tasks, many deep learning methods have been proposed to regress the parameters of such parametric human models from a single~\cite{pymaf2021, pymafx2022, PIXIE:2021} or multiple images~\cite{dong2021shape, zhang2021direct, burov2021dynamic}. 
However, these methods can only reconstruct a minimally-clothed human model without many details (\eg, hairs and clothes). 
Recently, methods based on implicit shape representation have reported encouraging performance, showing promising reconstruction with increased details in both single-view~\cite{saito2019pifu, zheng2021pamir, Huang_2020_CVPR, he2021arch++, xiu2022icon} and multi-view~\cite{hong2021stereopifu, shao2022doublefield, shao2022diffustereo, huang2018deep} settings.

Despite the stunning results reported by the above methods, their reconstructions remain far from perfect, restricting their practical applications. For instance, state-of-the-art (SOTA) single-view methods, such as PIFuHD~\cite{saito2020pifuhd}, PaMIR~\cite{zheng2021pamir}, and ARCH++~\cite{he2021arch++}, struggle with many self-occluding non-frontal human poses that widely occur in the real world. ICON~\cite{xiu2022icon} can handle these cases but the reconstructions contain serious artifacts (see \cref{fig:ours-sota}). 
On the other hand, many multi-view methods depend on calibrated cameras (\eg,~\cite{hong2021stereopifu, shao2022diffustereo, shao2022doublefield}) which are tedious to obtain in practice. Hence, how to carry out multi-view reconstruction with uncalibrated cameras is an important topic to study. Meanwhile, effective multi-view feature fusion is another key factor for robust mutli-view reconstruction. Prior techniques for multi-view feature fusion include average pooling~\cite{saito2019pifu}, SMPL-visibility~\cite{xiu2022icon}, and attention-based mechanism~\cite{zheng2021deepmulticap}. However, the reconstruction results based on these fusion techniques still contain notable artifacts in many cases. This indicates that more efforts are needed to derive a better multi-view feature fusion for more robust reconstruction.

In this paper, to extract more clothed human details flexibly and robustly from a single RGB image or uncalibrated multi-view RGB images, we present a novel framework, named \Netname, that employs the parametric model SMPL-X~\cite{SMPL-X:2019} as a 3D prior and combines the merits of both implicit and explicit representations. 
\Netname takes a single image or multi-view images as input to predict the occupancy for each 3D location in the space representing the human model. To reconstruct high-frequency details such as hairs and clothes, we introduce a self-evolved signed distance field module that learns to deform the signed distance field (SDF) derived from the fitted SMPL-X model using the input images. The resulting SDF can reflect more accurate geometry details than SMPL-X, which inherently deviates from the actual clothed human model. The SDF refined by our \Netname module is further encoded to allow for 3D reconstruction with better geometry details. 

Besides reconstructing a faithful 3D clothed human model from a single image,
our \Netname framework can also work with uncalibrated multi-view images to generate clothed human avatar with enhanced appearance (see \cref{fig:teaser}). To this end, we first propose a self-calibration method by fitting a shared SMPL-X model across multi-view images and projecting the shared model to different images based on the optimized rigid body motion for each input image. Further, we propose an occlusion-aware feature fusing strategy by probing the visibility of each 3D point under different views through ray-tracing, leveraging the SMPL-X model, such that features from visible views will contribute more to the fused feature while those from invisible views will be suppressed. 

The contributions are summarized as follows:
\begin{itemize}[leftmargin=*]
    \item We propose a flexible framework that, by leveraging the SMPL-X model as a shape prior, can take an arbitrary number of uncalibrated images to perform high-fidelity clothed human reconstruction. At the core of our framework is a self-evolved signed distance field (\Netname) module for recovering faithful geometry details.
    \item For uncalibrated multi-view reconstruction, we propose a simple self-calibration method through leveraging the SMPL-X model, as well as an occlusion-aware feature fusion strategy which takes into account the visibility of a space point under different views via ray-tracing.
    \item We thoroughly evaluate our framework on public benchmarks. SeSDF exhibits superior performances over current state-of-the-arts both qualitatively and quantitatively.
\end{itemize}

\section{Related Work}
\paragraph{Parametric 3D human model}
Parametric 3D human reconstruction~\cite{SMPL:2015, SMPL-X:2019, guan2009estimating, osman2020star, kanazawa2018end} paves the path for 3D human model. SMPL~\cite{SMPL:2015,pymaf2021} was built on convolutional networks and statistics collected from a wide range of human populations. SMPL-X~\cite{SMPL-X:2019, pymafx2022} extended SMPL with more body joints, keypoints, and expression parameters. 
Later substantial supervision strategies were proposed to regress the model parameters. Examples include semantic segmentation~\cite{omran2018neural, xu2019denserac}, human body silhouette~\cite{pavlakos2018learning}, joints~\cite{PIXIE:2021}, motion dynamics~\cite{kanazawa2018end, li2021hybrik}, texture consistency~\cite{pavlakos2019texturepose}, and attention-regressor~\cite{Kocabas_PARE_2021}. Multi-view~\cite{zhang2020object,dong2021shape, zhang2021direct, burov2021dynamic} and multi-person~\cite{ROMP, BEV} setups were also sought for various scenarios.
Complex human clothing and hair topology were also the goals for later parametric methods~\cite{joo2018total, xiang2019monocular, alldieck2018video, alldieck2019learning, zhu2019detailed}. 
However, due to the minimal-clothing assumption of the parametric models, these methods can retain little geometry details such as hairs and clothes in their reconstructions. 

\paragraph{Single-view (implicit) human reconstruction}
Many methods have been proposed in the deep learning era to reconstruct a 3D human model from a single image. For example,
voxel-based methods~\cite{gilbert2018volumetric, varol2018bodynet, Zheng2019DeepHuman, Trumble_2018_ECCV, jackson20183d, Zheng2019DeepHuman} reconstruct a volumetric representation of the human model. Two-stage inference~\cite{smith2019facsimile}, visual hull~\cite{natsume2019siclope} and front-back depth method~\cite{gabeur2019moulding} were then proposed to improve the reconstruction performance and efficiency. However, the high computational cost of these methods limits their real-world applications.
Implicit function methods have recently been introduced to predict either the occupancy field~\cite{mescheder2019occupancy, chibane20ifnet} or signed distance field~\cite{park2019deepsdf, chabra2020deep, jiang2020sdfdiff} using neural networks. PIFu~\cite{saito2019pifu} presents promising results with a pixel-aligned implicit function. Follow-up works exploited 3D features to solve the depth-ambiguity problem~\cite{zheng2021pamir}, achieve animatable model~\cite{Huang_2020_CVPR, he2021arch++}, or improve the reconstruction details~\cite{cao2022jiff}. PIFuHD~\cite{saito2020pifuhd} and FITE~\cite{lin2022fite} apply a multi-stage method to improve clothing topology in the reconstruction. Geo-PIFu~\cite{he2020geopifu} proposes to compute 3D features directly from images. 
ICON~\cite{xiu2022icon} exploits the SMPL-based signed distance filed as guidance to help produce a robust reconstruction, with a smpl-visibility based method to choose features from front or back sides. PHORHUM~\cite{alldieck2022phorhum} explores shading and illumination information to infer better geometry and texture.
Encouraging improvements have been reported by these methods. However, their performance on non-frontal human images are still far from being robust and satisfactory. 

\paragraph{Multi-view implicit human reconstruction}
Multi-view reconstruction is conceptually a more natural choice as information from different views can compensate each other for better reconstruction. Under a calibrated setting, PIFu~\cite{saito2019pifu} proposes to fuse features from multi-views by average pooling after being embedded using an MLP. StereoPIFu~\cite{hong2021stereopifu} takes as input a pair of stereo images, and performs depth-aware reconstruction. DiffuStereo~\cite{shao2022diffustereo} employs diffusion models for multi-view stereo human reconstruction. DoubleField~\cite{shao2022doublefield} combines the surface field and radiance fields to improve both geometry and texture inference. Effi-MVS~\cite{wang2022efficient} explores the dynamic cost volume for reduced computational cost. DeepMultiCap~\cite{zheng2021deepmulticap} handles the multi-person reconstruction with multi-view images.
Neural-Human-Performer~\cite{kwon2021neural} proposes a multi-view transformer to fuse features for multi-view reconstruction. 
Despite promising results have been achieved, these methods require a large number of views and camera calibration. MVP-Human~\cite{zhu2022mvp} makes use of the deformation fields and handles multi-view inputs under an uncalibrated setting, but the reconstruction quality lags far behind the calibrated methods. For calibrated methods, their models still produce undesirable artifacts for many self-occluding non-frontal poses and miss many geometry details. 

\vspace{-0.5em}
\section{Method}

\begin{figure*}[h]
  \centering
   \includegraphics[width=1\linewidth]{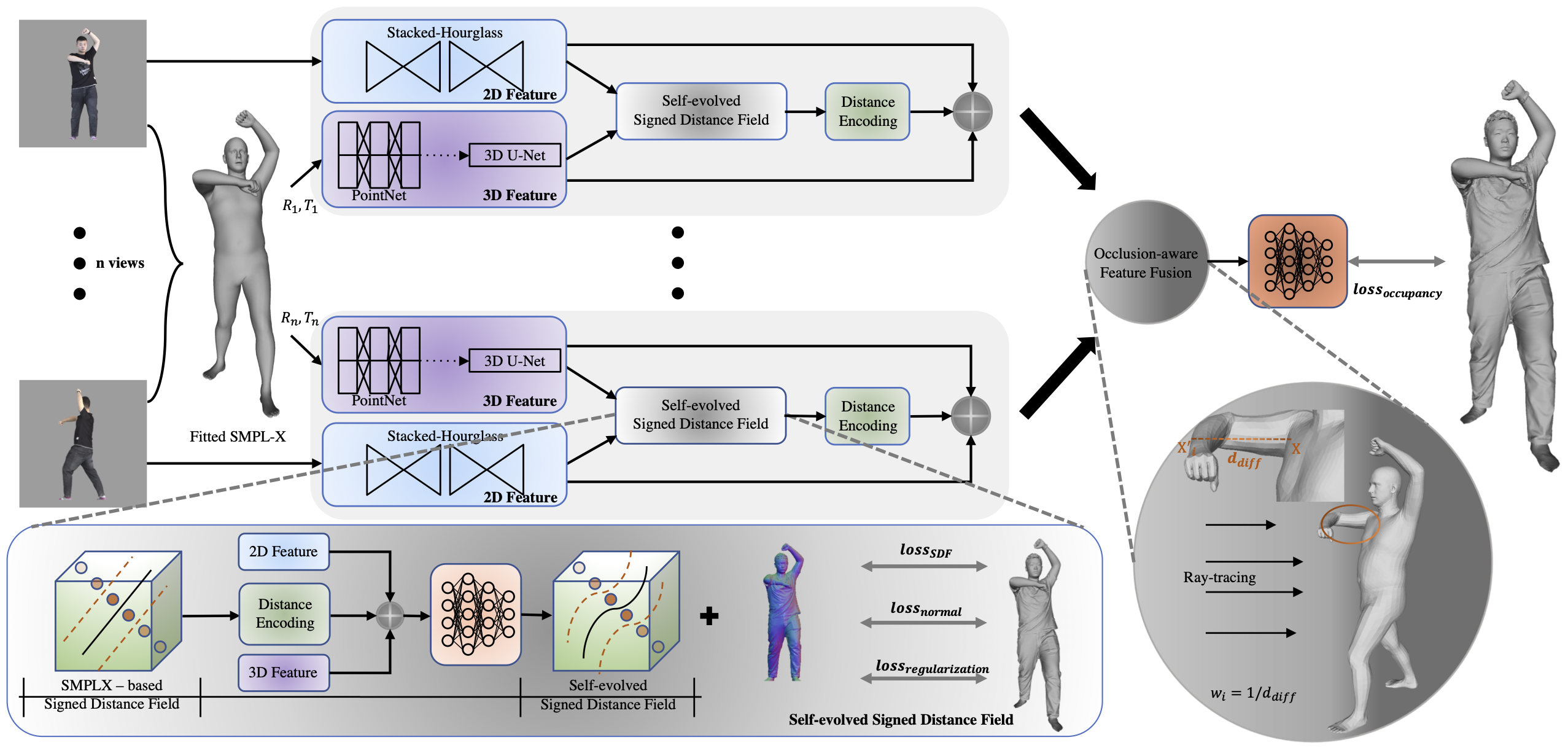}

   \caption{Overview of our proposed \Netname framework. It takes a single image or uncalibrated multi-view images as input to predict the occupancy for each 3D location in the space for shape reconstruction. We first fit a SMPL-X model to the input image(s) which serves as a shape prior and provides space-aligned 3D features. At the core of our method is our proposed self-evolved signed distance field (\Netname) module, which learns a refined  signed distance field (SDF) to better reflect the actual clothed human subject rather than the minimally-clothed SMPL-X model. \Netname module takes  pixel-aligned image features, space-aligned 3D features, SDF with distance encoding, and SMPL-X vertex normal as input to predict the refined SDF and normal value. For multi-view reconstruction, we propose a SMPL-X based self-calibration method to eliminate the need for manual calibration and an occlusion-aware feature fusion strategy to effectively accumulate features from different views.
   }
   \label{fig:ssdpipeline}
\end{figure*}

We propose an implicit function based framework for reconstructing 3D clothed human models from a single image or uncalibrated multi-view images (see~\cref{fig:ssdpipeline}). At the core of our framework are a self-evolved signed distance field (\Netname) module and an occlusion-aware feature fusion module. 
Next, we will first introduce the preliminary for \Netname in~\cref{sec:pre}. We will then describe \Netname for single-view reconstruction in~\cref{sec:single}, followed by uncalibrated multi-view reconstruction in~\cref{sec:multi}.

\subsection{Preliminary}
\label{sec:pre}
\paragraph{Surface by implicit function} 
Implicit functions~\cite{dontchev2009implicit} have been shown to be effective for 3D reconstruction from images or videos. Typically, an implicit function is implemented as an MLP, which takes the 3D coordinate $X$ and an optionally associated feature $F$ of a 3D point as input, and outputs the corresponding occupancy as a scalar value. Commonly employed features include pixel-aligned image features~\cite{saito2019pifu, saito2020pifuhd} and space-aligned 3D features~\cite{Huang_2020_CVPR, zheng2021pamir, cao2022jiff}. Formally, the 
implicit function can be written as 
\begin{equation}\label{implicitFunction}
    f_v(X, F) \mapsto [0, 1] \in \mathbb{R}.
\end{equation}

\paragraph{SMPL-X}
The parametric human model SMPL-X represents the 3D shape by incorporating body vertices, joints, face and hands landmarks, and expression parameters. Formally, by assembling pose parameters $\theta$, shape parameters $\beta$, and facial expression parameters $\phi$, SMPL-X human model can be expressed by a function $M(\cdot)$ as
\begin{eqnarray*}\label{SMPLX}
    \small
    M(\beta, \theta, \phi)&=&W(T_p(\beta, \theta, \phi), J(\beta), \theta, \mathcal{W}), \\ 
    T_p(\beta, \theta, \phi)&=&\Bar{T} + B_S(\beta; \mathcal{S}) + B_E(\phi; \mathcal{E}) + B_p(\theta; \mathcal{P}),
\end{eqnarray*}
where $W(\cdot)$ is the linear blend skinning function, $J(\beta)$ is the joint location, and $\mathcal{W}$ denotes the blend weights. $T_p(\cdot)$ poses the SMPL-X canonical model $\Bar{T}$ by shape blend shape function $B_S$, expression blend shape function $B_E$, and pose blend shape function $B_P$. $\mathcal{S}$ and $\mathcal{P}$ are the principal components of vertex displacements, and $\mathcal{E}$ represents the principal components 
capturing variations of facial expressions.

\paragraph{SMPL-X optimization}
Given a human image, we can fit the SMPL-X model parameters by~\cite{PIXIE:2021}. However, such fitted models often suffer from misalignment. We use the well-aligned SMPL-X models provided in the training data to provide shape prior to train our framework. At test time, we do not have such well-aligned models. Hence, to avoid errors and artifacts caused by the misalignment during testing, we optimize the SMPL-X parameters by projecting the SMPL-X model onto the input image and maximizing the Intersection over Union (IoU) between the projection and body mask~\cite{cheng2022generalizable, xiu2022icon}, and minimizing the 2D keypoint distances.
Please refer to the supplementary for details.

\subsection{Single-view reconstruction}
\label{sec:single}
Given an input image $\boldsymbol{I}$ and an optimized SMPL-X mesh $\boldsymbol{S}$, \Netname incorporates pixel-aligned image features $F_{2D}$ extracted from $\boldsymbol{I}$, space-aligned 3D features $F_{3D}$ extracted from $\boldsymbol{S}$, and the self-evolved signed distance field $d$ to predict occupancy using an MLP. By applying marching cube~\cite{li2018differentiable} on the occupancy predictions in the 3D space, \Netname robustly reconstructs a 3D clothed human model with faithful details, including the clothing topology. Given a 3D point $X$, the implicit function for single-view reconstruction can be formulated as
\begin{equation}\label{ssdfunction}
    \small
    f_o(F_{2D}(X), F_{3D}(X), \mathcal{D}(d(X)), \mathbf{n}(X), Z(X))  \mapsto [0, 1],
\end{equation}
where $\mathcal{D}(\cdot)$ and $Z(\cdot)$ denote distance encoding and depth value respectively, and $d(\cdot)$ and $\mathbf{n}(\cdot)$ are the signed distance and normal output by our \Netname module.

\paragraph{Pixel-aligned image feature} We apply a stacked-hourglass encoder~\cite{newell2016stacked, saito2019pifu} to the input image $I \in \mathbb{R}^{512\times512\times3}$ to produce an image feature map $E_I(I)\in \mathbb{R}^{256\times128\times128}$. We then obtain the pixel-aligned image feature for the projection of a 3D point using bi-linear interpolation.

\paragraph{Space-aligned 3D feature} We apply PointNet~\cite{qi2017pointnet} and 3D-UNet~\cite{cciccek20163d} to extract 3D features~\cite{cao2022jiff, Peng2020ECCV} from the optimized SMPL-X mesh. Our 3D encoder first processes the whole 10,475 vertices of the mesh to generate hierarchical 3D point features, which are transformed into a 3D feature volume via average pooling. 3D-UNet is then applied to output the final 3D feature volume $E_S(\boldsymbol{S}) \in \mathbb{R}^{128\times64\times64\times64}$. Finally we obtain the space-aligned 3D feature for a 3D point via tri-linear interpolation.

\paragraph{Self-evolved signed distance field}
The signed distance field (SDF) derived from the SMPL-X model~\cite{xiu2022icon} provides a strong prior for estimating the occupancy. However, the SMPL-X model does not model geometric details of the human, such as the cloth topology, and thus the resulting SDF will unavoidably deviate from the true SDF of the actual clothed human model. Directly employing this SDF for occupancy prediction will therefore induce error in the reconstruction.
To remedy this problem and recover plausible geometry details in the reconstruction, we propose a novel self-evolved signed distance filed (\Netname) module. This module is trained to evolve the SDF derived from the SMPL-X model to better reflect the actual clothed human model. Instead of predicting only the signed distance $d(\cdot) \in \mathbb{R}$, our \Netname module further predict the normal $\mathbf{n}(\cdot) \in \mathbb{R}^3$ which provides additional surface orientation information to guide the implicit function to better predict occupancy and capture fine shape details.
Concretely, \Netname module is implemented as an MLP and formulated as
\begin{equation}\label{ssdmodule}
    f_{sd}(F_{2D}(X), F_{3D}(X), \mathcal{D}(d'(X)), \mathbf{n}'(X))  \mapsto (d, \mathbf{n}), 
\end{equation}
where $d'(\cdot) \in \mathbb{R}$ denotes the SDF derived from the SMPL-X model and $\mathbf{n}'(\cdot) \in \mathbb{R}^3$ represents the normal obtained from the closest SMPL-X vertex to $X$. 

\paragraph{Distance encoding}
Inspired by the positional encoding adopted in NeRF~\cite{mildenhall2021nerf}, we adopt a distance encoding that maps the signed distance $d$ to a higher dimensional space using high frequency functions:
\begin{equation}\label{distenc}
    \small
    \mathcal{D}(d) = (d, \sin(2^0\pi d), \cos(2^0\pi d), ..., \sin(2^{L}\pi d), \cos(2^{L}\pi d)).
\end{equation}
In our experiments, we set $L = 5$. The encoded signed distance $\mathcal{D}(d)$ is concatenated with the predicted normal $\mathbf{n}(\cdot)$, 2D feature $F_{2D}(\cdot)$ and 3D feature $F_{3D}(\cdot)$, and fed to the implicit function $f_o(\cdot)$ for occupancy prediction.

\paragraph{Training objectives}
To consistently regress \Netname and simulate realistic cloth topology, we consider two groups of sampling points, namely surface samples $G_s$, and occupancy-target samples $G_o$. Points in $G_s$ are uniformly sampled on the mesh surface, while points in $G_o$ are sampled following the strategy proposed in PIFu~\cite{saito2019pifu}.

For points in $G_s$, we adopt the following loss function
\begin{equation}
    \small
    \mathcal{L}_{s} = \frac{1}{|G_s|}\sum_{X\in G_s} \lambda_d |d(X)| + \lambda_n ||\mathbf{n}(X) - \mathbf{n}_{gt}(X)||,
\end{equation}
where $\mathbf{n}_{gt}(X)$ denotes the ground-truth normal at the surface point $X$. For points in $G_o$, we adopt the following loss function
\begin{equation}
    \small
    \mathcal{L}_o = \frac{1}{|G_o|}\sum_{X\in G_o} BCE(\mathcal{O}(X) - \mathcal{O}_{gt}(X)),
\end{equation}
where $\mathcal{O}(X)$ denotes the predicted occupancy at $X$ computed using \cref{ssdmodule} and \cref{ssdfunction}, $\mathcal{O}_{gt}(X)$ is the ground-truth occupancy at $X$, and $BCE(\cdot)$ denotes the binary cross entropy.
Following IGR~\cite{gropp2020implicit}, we add an eikonal normal regularization term to smooth the surface by: 
\begin{equation}
    \mathcal{L}_r = \frac{1}{|G_o|}\sum_{X\in G_o} (||\mathbf{n}(X)||-1)^2.
\end{equation}

The overall loss function is given by
\begin{equation}\label{ssdloss}
    \mathcal{L}_{SeSDF} = \lambda_s \mathcal{L}_s + \lambda_o \mathcal{L}_o + \lambda_r \mathcal{L}_r,
\end{equation}
where $\lambda_s$, $\lambda_o$, and $\lambda_r$ are weights to balance the effect of the respective loss.

\subsection{Uncalibrated multi-view reconstruction}
\label{sec:multi}
To allow our \Netname framework work with uncalibrated multi-view images, we present a simple self-calibration method to estimate the relative camera pose for each image through fitting a SMPL-X model to all the input images. In addition, we propose an occlusion-aware feature fusion strategy to fuse features from different views effectively. Ideally, we would like to fuse only features from non-occluded views as features from occluded views are just noise which at best provide zero contribution to occupancy prediction but may also seriously degrade the fused feature in the worse case.

Our multi-view framework is illustrated in \cref{fig:ssdpipeline}.
In principle, it can work with an arbitrary number of views and the number of views used in training and testing do not even need to be the same. 

\begin{table*}[t]
\centering
\caption{Qualitative comparison on the single-view setting. We report Chamfer Distance and P2S metrics (numbers the lower the better).
}
\label{tab:quantitative_single}
\resizebox{1.\textwidth}{!}{
\begin{tabular}{l||ccccc||cccc}

\toprule[1pt]
{\multirow{3}*{Method}} & \multicolumn{5}{c||}{\multirow{2}*{Feature Included}} & \multicolumn{4}{c}{Quantitative Number} \\
{} & \multicolumn{5}{c||}{} & \multicolumn{2}{c}{THuman2.0}  & \multicolumn{2}{c}{CAPE} \\
\cmidrule[0.5pt](rl){2-6}
\cmidrule[0.5pt](rl){7-8}
\cmidrule[0.5pt](rl){9-10}
{} & {2D Feature} & {Normal Information} & {3D Feature} & {SMPLX Signed Distance} &{self-evolved Signed Distance} & {Chamfer Distance$\downarrow$}& {P2S$\downarrow$} & {Chmafer Distance$\downarrow$} & {P2S$\downarrow$} \\
\midrule[1pt]
PIFu~\cite{saito2019pifu}   & $\checkmark$ & & & & & 2.403 & 2.388 & 3.059 & 2.818\\         
PIFuHD~\cite{saito2020pifuhd}  & $\checkmark$ & $\checkmark$ & & & & 2.008 & 1.965 & 2.571 & 2.427\\
PaMIR~\cite{zheng2021pamir}  & $\checkmark$ & & $\checkmark$ & & & 1.478 & 1.451 & 1.682 & 1.438 \\
ICON~\cite{xiu2022icon}  & $\checkmark$ & $\checkmark$ & & $\checkmark$ & & 1.301 & 1.259 & 1.533 & 1.431\\
\midrule[1pt]
Ours w/o DE  & $\checkmark$ & $\checkmark$ & $\checkmark$ & & $\checkmark$ & 1.099 & 1.034 & 1.377 & 1.261\\
Ours w/o SeSDF & $\checkmark$ & $\checkmark$ & $\checkmark$ & $\checkmark$ & & 1.245 & 1.200 & 1.473 & 1.390 \\
Ours  & $\checkmark$ & $\checkmark$ & $\checkmark$ & & $\checkmark$ & \textbf{1.027} & \textbf{0.971} & \textbf{1.303} & \textbf{1.219}\\
\midrule[1pt]
\end{tabular}
}
\end{table*}
\paragraph{Self-calibration via SMPL-X model fitting} 
Given $n$ uncalibrated multi-view images, we first fit a SMPL-X model to each image independently using the method described in \cref{sec:pre}. We then take the averages of the shape, pose and expression parameters of these $n$ fitted models and use them to initialize a shared SMPL-X model. Next, we optimize the shape, pose and expression parameters of this shared SMPL-X model and the global orientation parameters for the $n$ views simultaneously by maximizing the IoU between the projection and body mask, and minimizing the 2d keypoint distances on each of the $n$ images. After optimization, the global orientation parameters for the $n$ views then give us the rigid body motion that transforms a point from the SMPL-X model centered coordinate system to the camera centered coordinate system of each view.

To perform multi-view reconstruction, we can sample 3D points in the model centered coordinate system, transform them into the respective camera coordinate systems, and compute 2D/3D features and the self-evolved signed distance from each view. Finally, these features from $n$ different views are fused by our occlusion-aware feature fusion strategy (to be introduced next) and fed to the implicit function $f_o(\cdot)$ to predict the occupancy value.

\paragraph{Occlusion-aware feature fusion} 
Conceptually, features obtained from different views would give rise to predictions of different quality. For instance, image feature for a 3D point extracted from a frontal non-occluded view would be expected to give the best prediction, whereas image feature for the same point but extracted from an occluded view or a lateral view would be expected to give poor prediction. Hence, average pooling would in principle not be the best way for feature fusion. Researchers therefore explored different fusion strategies, such as SMPL-visibility based~\cite{xiu2022icon} and attention based~\cite{zheng2021deepmulticap} approaches, but all of these still produce undesirable artifacts in their reconstructions (see \cref{fig:ablation_ff}). To lessen the effect of features extracted from non-optimal views, we propose a novel occlusion-aware feature fusion strategy based on ray tracing~\cite{glassner1989introduction} with the fitted SMPL-X model. Given a 3D point $X$, we cast a ray from $X$ to the $i$-th view through orthogonal projection and find the intersection point $X_i'$ with the SMPL-X model that is the closest to the image (see \cref{fig:ssdpipeline}). We then compute a weight for the feature $F_{i}(X)$ extracted from the $i$-th view by 
\begin{equation}
    w_i(X) = \frac{1}{|Z_i(X) - Z_i(X_i')|}.
\end{equation}
Finally, we fuse all features from the $n$ views by
\begin{equation}
    F_{fused}(X) = \frac{\sum_{i=1}^{n} w_i(X) F_i(X)}{\sum_{i = 1}^n w_i(X)}.
\end{equation}

\begin{table}[t]
\centering
\caption{Quantitative comparison on the multi-view setting. 
}
\label{tab:quantitative_sparse}
\resizebox{0.48\textwidth}{!}{
\Large

\begin{tabular}{l||cccc}

\toprule[1pt]
{} & \multicolumn{2}{c}{THuman2.0}  & \multicolumn{2}{c}{CAPE} \\
\cmidrule[1pt](rl){2-3}
\cmidrule[1pt](rl){4-5}
{Method} & {Chamfer Distance$\downarrow$} & {P2S$\downarrow$} & {Chmafer Distance$\downarrow$} & {P2S$\downarrow$}\\
\midrule[1pt]
multi-view PIFu~\cite{saito2019pifu} & 0.864 & 0.838 & 0.920 & 0.876\\     
multi-view PIFuHD~\cite{saito2020pifuhd} & 0.831 & 0.804 & 0.883 & 0.849 \\         
multi-view PaMIR~\cite{zheng2021pamir} & 0.807 & 0.762 & 0.858 & 0.819 \\
\cmidrule[1pt](rl){1-5}
Ours w/o DE & 0.795 & 0.749 & 0.849 & 0.811\\
Ours w/o SeSDF & 0.802 & 0.755 & 0.854 & 0.815\\
Ours & \textbf{0.791} & \textbf{0.742} & \textbf{0.841} & \textbf{0.800}\\
\cmidrule[1pt](rl){1-5}
Ours w/ Attention & 0.794 & 0.743 & 0.845 & 0.803\\
Ours w/ SMPLX-Vis & 0.841 & 0.832 & 0.925 & 0.917 \\
Ours w/ Normal-Fusion & 0.801 & 0.754 & 0.850 & 0.812 \\
\bottomrule[1pt]
\end{tabular}
}
\end{table}

\section{Experiments}
We evaluate the performance of our \Netname both qualitatively and quantitatively, 
\begin{figure}[h]
  \centering
   \includegraphics[width=1\linewidth]{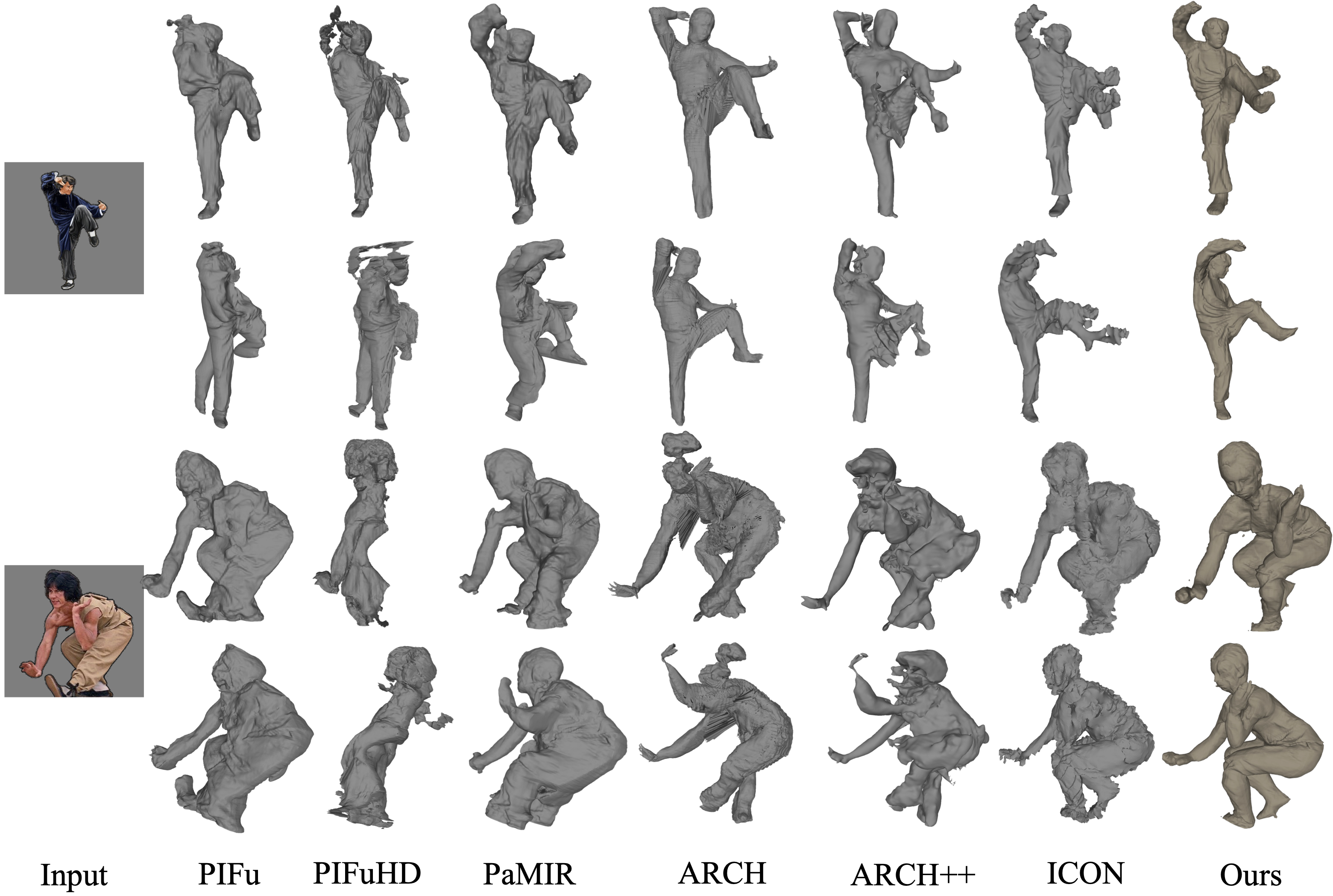}
  \caption{
   Qualitative comparison with SOTA methods on real-world images. Our method can reconstruct fine details and work robustly under challenge poses. Best viewed in PDF with zoom.
  }
   \label{fig:ours-sota}
   
\end{figure} 
and provide comparisons with other SOTA methods under both single-view and multi-view setups.
\subsection{Implementation details}
For each training subject, we sample (a) 5,000 surface points as $G_s$, which are evenly distributed on the mesh surface; 
(b) 5,000 occupancy-target points as $G_o$ in the 3D space, for which we follow the sampling strategy in~\cite{saito2019pifu}. In particular,
15/16 points in $G_o$ are evenly sampled on the mesh surface. Gaussian perturbation is then applied to them along the surface normal direction. The rest 1/16 points are randomly sampled within the predefined 3D space where the mesh lies in. 
We employ Rembg~\cite{rembg} to segment out the background for real-world images and use Kaolin~\cite{kaolin} to calculate the SDF from the SMPL-X model. We apply PIXIE~\cite{PIXIE:2021} to estimate the SMPL-X model for the segmented human subject. To compute the image features, we implement a stacked-hourglass image encoder with bi-linear interpolation to extract pixel-aligned image feature $F_{2D}(\cdot) \in \mathbb{R}^{256}$. 
We follow \cite{cao2022jiff, Peng2020ECCV} to employ PointNet with 3D-UNet and tri-linear interpolation to extract space-aligned 3D features $F_{3D}(\cdot) \in \mathbb{R}^{128}$. 
Distance encoding maps the signed distance to a higher dimensional space $\mathcal{D}(d) \in \mathbb{R}^{13}$ using the high-frequency functions as defined in \cref{distenc}.
We implement our framework in PyTorch~\cite{paszke2017pytorch}.
Our \Netname framework is trained separately for 12 epochs with the learning rate starting from 1e-4 and updated by the factor 0.1 for every 4 epochs, for both single-view and multi-view experiments. Code for SeSDF will be made publicly available. Project page: \url{https://yukangcao.github.io/SeSDF/}.

\subsection{Datasets and evaluation metrics}
\begin{figure*}[h]
  \centering
   \includegraphics[width=1\linewidth]{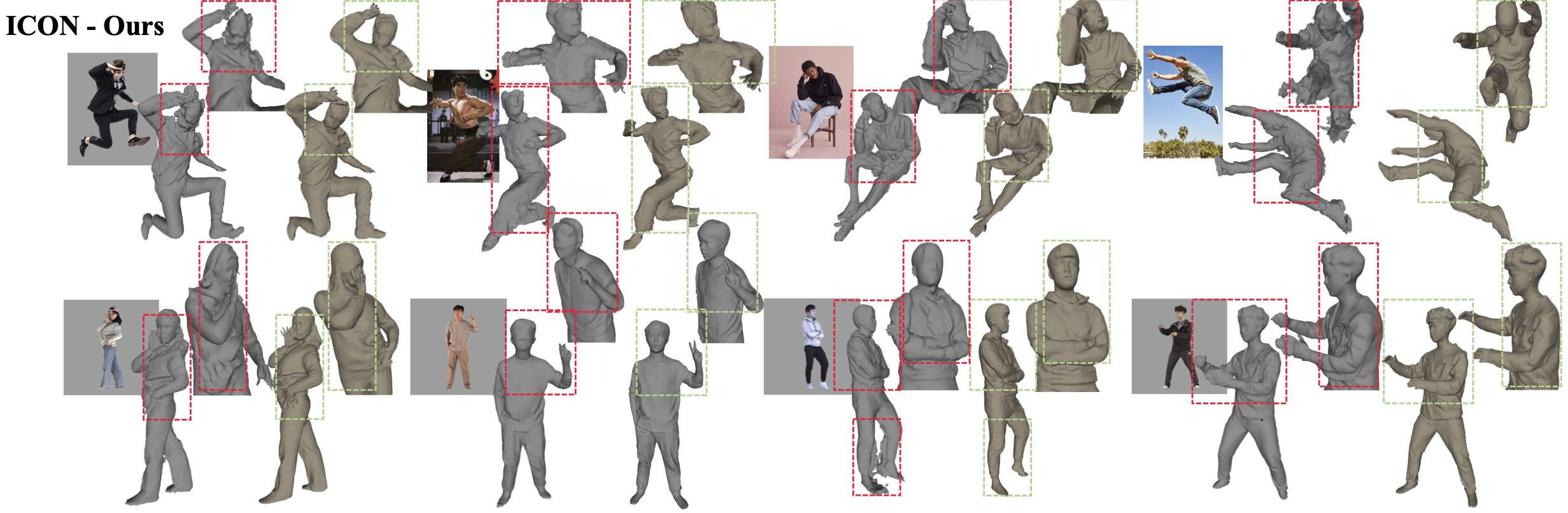}

  \caption{Comparison with ICON on real-world (top) and THUman2.0 data (bottom). Ours (SeSDF) can produce results with more fine geometry details. Best viewed in PDF with zoom.}
   \label{fig:ours-icon_sub}
\end{figure*}
We employ the THUman2.0 dataset~\cite{tao2021function4d} as our primary testbed. The dataset comprises 526 high-fidelity meshes captured via dense DSLR rigs. Each subject within the dataset is accompanied by its corresponding texture map and fitted SMPL-X parameters. We split the THUman2.0 dataset into a training set with 465 subjects, and a test set with 61 subjects. We render each human subject into 360 images, sequentially separated by $1^{\circ}$ around the subject, with different albedo to augment the dataset to carry out the experiments.

To further show the generalizability of \Netname, we also compare our method with others on the CAPE~\cite{CAPE:CVPR:20} test set, which contains 150 subjects. To conduct quantitative evaluations of different methods, we employ standard Point-to-Surface distance (P2S) and Chamfer distance metrics as evaluation criteria. We compare the reconstructed models from images captured at different angles, \ie, $0^{\circ}$, $45^{\circ}$, $90^{\circ}$, $135^{\circ}$, $180^{\circ}$, $225^{\circ}$, $270^{\circ}$, and $315^{\circ}$, against the ground truth under corresponding orientations. Besides, we use real-world images for extra qualitative evaluation.

\subsection{Comparison on single-view setting}
We first evaluate and compare the performance of single-view 3D human reconstruction between ours and SOTA methods. Specifically, we compare with PIFu~\cite{saito2019pifu}, PIFuHD~\cite{saito2020pifuhd}, PaMIR~\cite{zheng2021pamir}, ARCH~\cite{Huang_2020_CVPR}, ARCH++~\cite{he2021arch++} and ICON~\cite{xiu2022icon}. 
PHORHUM~\cite{alldieck2022phorhum} is not compared here due to the absence of publicly available code base. 
We retrain PIFu, PIFuHD, PaMIR, and ICON on the THUman2.0 dataset for a fair comparison, and obtain the reconstructions of ARCH and ARCH++ with the help of their authors. 

\paragraph{Quantitative comparison and ablation}
In~\cref{tab:quantitative_single}, we compare our method with the SOTA methods, \ie, PIFu, PIFuHD, PaMIR, and ICON on the THUman2.0 and CAPE datasets. As can be seen, for all metrics, our method significantly outperforms the SOTA by a large margin on both datasets. Among these SOTA methods, ICON also employs the SMPL-X model to extract SDF to guide learning, while our method consistently outperforms ICON across the board. 
We also validate the effectiveness of different components of our method. All components have positive effects on the reconstruction results, while the SeSDF module plays the most crucial role in the improvement. By comparing the last two rows, we can see that the SeSDF module brings much more improvement than the SDF derived from the SMPL-X model. By comparing the last row with all others, we can observe that our SeSDF brings the most improvement among all components. By comparing ``Ours'' and ``Ours w/o DE'', we can see that our proposed distance encoding is also effective in improving the performance. 

\paragraph{Qualitative comparison}
We present qualitative comparison in~\cref{fig:ours-sota} on real-world images with SOTA methods. Our proposed method has the capability to reconstruct fine geometry details and handle the inputs with complicated poses. 
Among the SOTA methods, ICON also employs SMPL-X to improve the robustness, though in a different way from our work. We provide a further comparison with ICON in~\cref{fig:ours-icon_sub} on THUman2.0 and real-world images.
As can be seen, our method notably outperforms ICON with more fine details in all cases.

\subsection{Comparison on multi-view setting}
For multi-view reconstruction, we compare with the SOTA methods PIFu, PIFuHD, and PaMIR which allow multi-view reconstruction. In the experiments, we use three input views unless stated otherwise. 

\paragraph{Quantitative comparison and ablation}
We report the comparison results in \cref{tab:quantitative_sparse}. 
To have a fair comparison with other methods which are calibrated, we also implemented our self-calibration method for all of them. We can see that our method outperforms other methods in all cases. Similar to the single-view experiments, by comparing ``Ours'' against ``Ours /wo DE'' and ``Ours /wo SeSDF'',  we can see that our \textit{distance encoding} and \textit{SeSDF} play critical roles in the performance.

\begin{figure*}[h]
  \centering
   \includegraphics[width=1\linewidth]{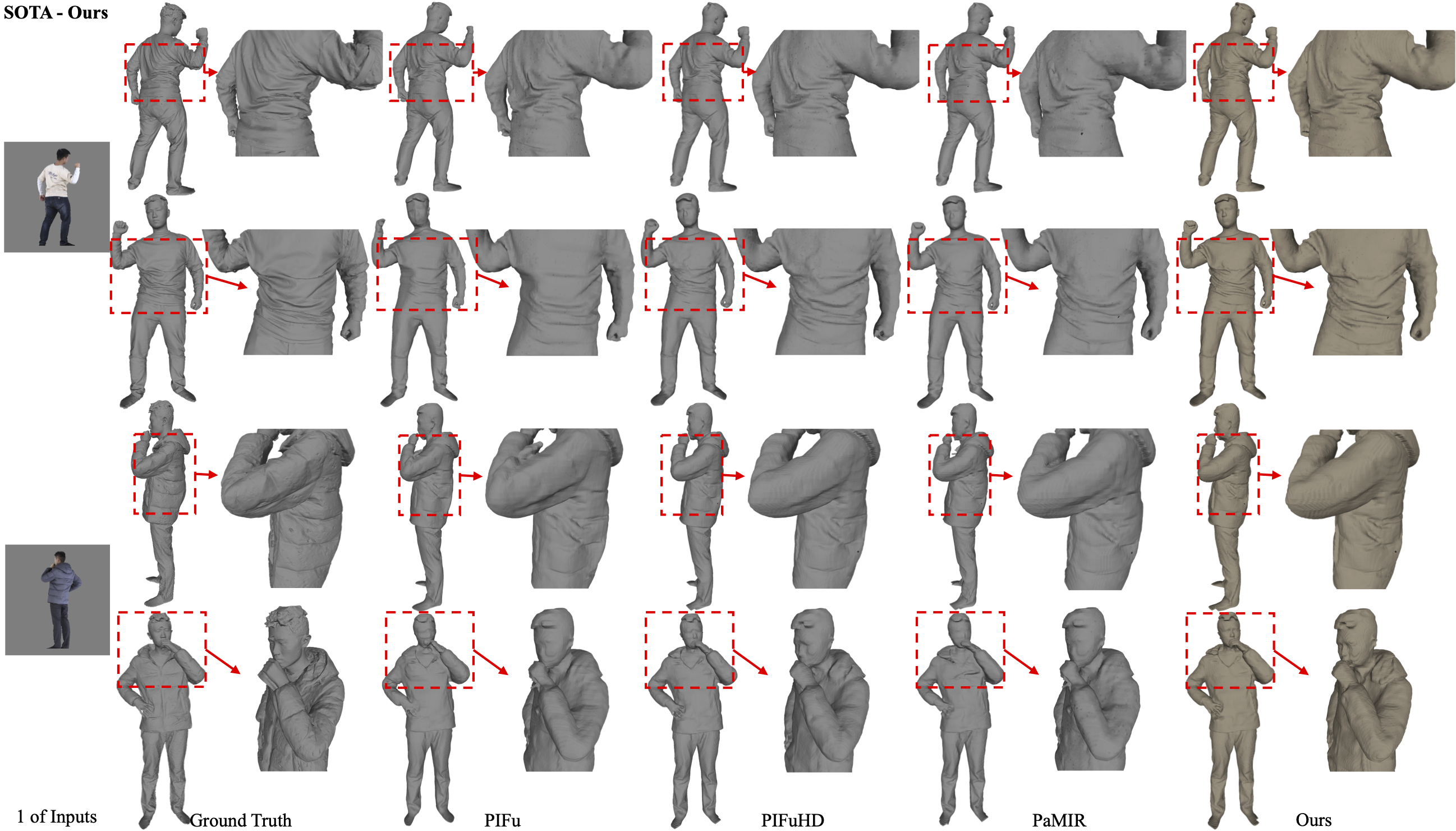}

  \caption{Comparison with SOTA methods on the multi-view setting. Ours (SeSDF) consistently retains more geometry details, \eg, clothing wrinkles, hairs, and faces.}
   \label{fig:sparseview_comparison}
\end{figure*}

\paragraph{Qualitative comparison}
We present qualitative comparison in~\cref{fig:sparseview_comparison} for multi-view reconstruction. The reconstruction results of our method retain more details, \eg, clothing wrinkles and hair topology, which are contained in the ground truth.
We provide more results with different numbers of views in the supplementary.
\vspace{-1.0em}
\subsubsection{Further analysis}

\begin{figure}[ht]
  \centering
   \includegraphics[width=1\linewidth]{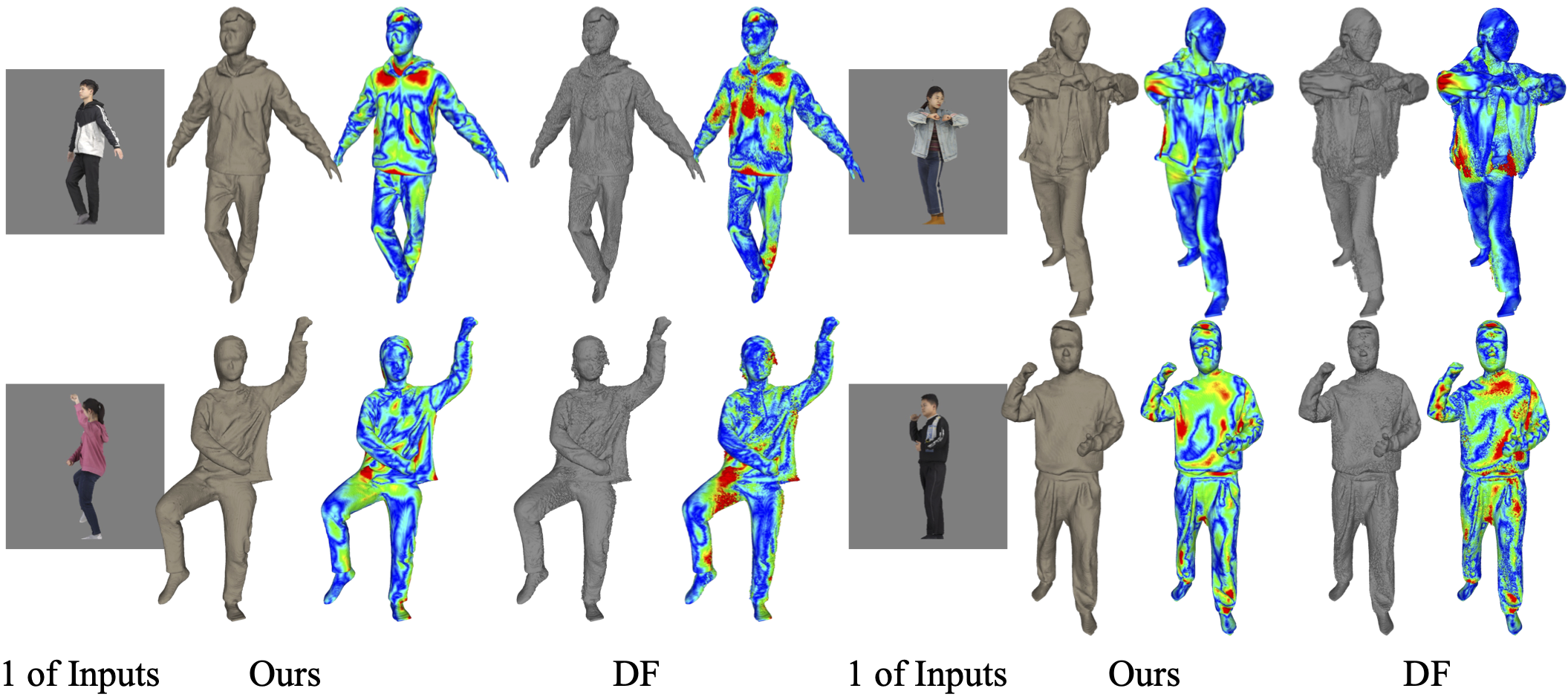}

   \caption{Comparing our self-calibration method (``Ours'') with an SMPL-X deformation field alternative (``DF''). Best viewed in PDF with zoom.}
   \label{fig:ablation_go}
\end{figure}

\paragraph{Self-calibration}
\label{sec:ablation_go}
\cite{zhu2022mvp} also introduced an alternative method for self-calibration of multi-view reconstruction by introducing a deformation field to deform points from the canonical space to the camera space, through SMPL linear blend skinning (LBS) weights. We also follow the paper to implement it for SMPL-X in our framework and compare it with our method in~\cref{fig:ablation_go}. We observe wide-spreading artifacts across the whole body of the reconstruction using the deformation field for the uncalibrated setting. 
The deformation field is error-prone because the LBS weights are computed to deform the parametric human model, which inherently lacks the fine details of the actual human subject, limiting its final performance and introducing the artifacts for reconstruction.

\begin{figure}[ht]
  \centering
   \includegraphics[width=1\linewidth]{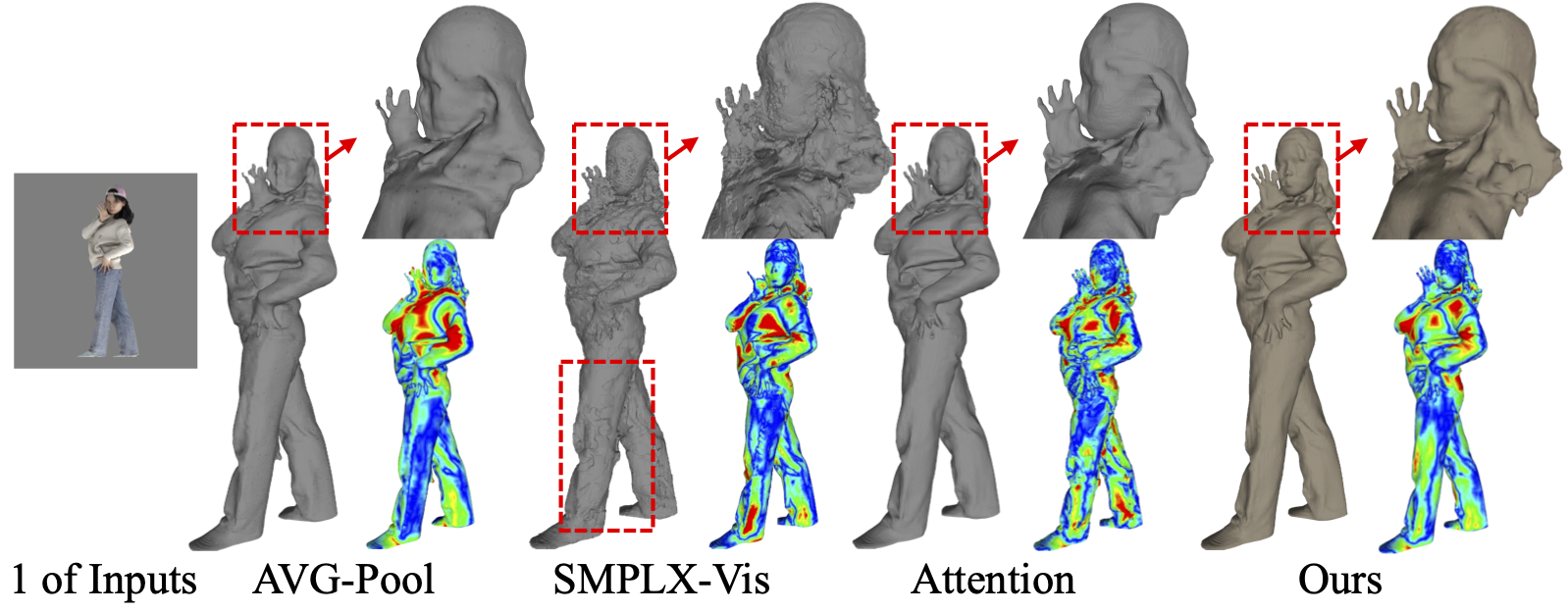}

   \caption{Comparing our occlusion-aware feature fusion with other alternative fusion strategies. }
   \label{fig:ablation_ff}
	\vspace{0.5em}
\end{figure}

\paragraph{Feature fusion}
We compare our occlusion-aware feature fusion strategy with other alternatives, namely (1) AVG-Pool~\cite{saito2019pifu}, (2) SMPLX-visibility~\cite{xiu2022icon}, and (3) Attention-based method~\cite{kwon2021neural} in~\cref{fig:ablation_ff}. As can be seen, our occlusion-aware strategy achieves better performance than the others. See supplementary for more analysis.

\section{Conclusion}
In this paper, we have presented a new framework for single- and multi-view clothed human reconstruction under an uncalibrated setting. Our framework can take an arbitrary number of input images to reconstruct faithful 3D human models. We have proposed a self-evolved signed distance (\Netname) module to recover more geometry details, a method for self-calibrating multi-view images via SMPL-X model fitting, 
and an effective multi-view occlusion-aware feature fusion strategy to aggregate useful features across views for robust reconstruction. Our method notably outperforms previous SOTA methods in both terms of qualitative and quantitative results.

\paragraph{Limitation} 
Though our SeSDF can deal with various poses and retain fine clothing topology, it still has difficulties when handling extremely loose clothes, \eg, dresses, which deviate significantly from the parametric prior.

\paragraph{Potential Social Impact} Our method takes a step towards practical tools for real-world application, as it does not require expensive equipment to get the 3D models. However, despite its intended positive usage,  there might be potential risks of falsifying human avatars, which unfortunately compromise personal privacy. Transparency and authentication will be helpful to prevent these vicious uses.

\paragraph{Acknowledgements}
This work is partially supported by Hong Kong Research Grant Council - Early Career Scheme (Grant No. 27208022) and HKU Seed Fund for Basic Research. 
We appreciate the very helpful discussions with Dr. Guanying Chen. 
We also thank Yuliang Xiu, Yuanlu Xu for ARCH and ARCH++ results.


\clearpage

\renewcommand\thesection{\Alph{section}}
\setcounter{section}{0}
\onecolumn
\section{Implementation details}

\subsection{SMPL-X optimization}
\begin{figure}[h]
  \centering
   \includegraphics[width=0.8\linewidth]{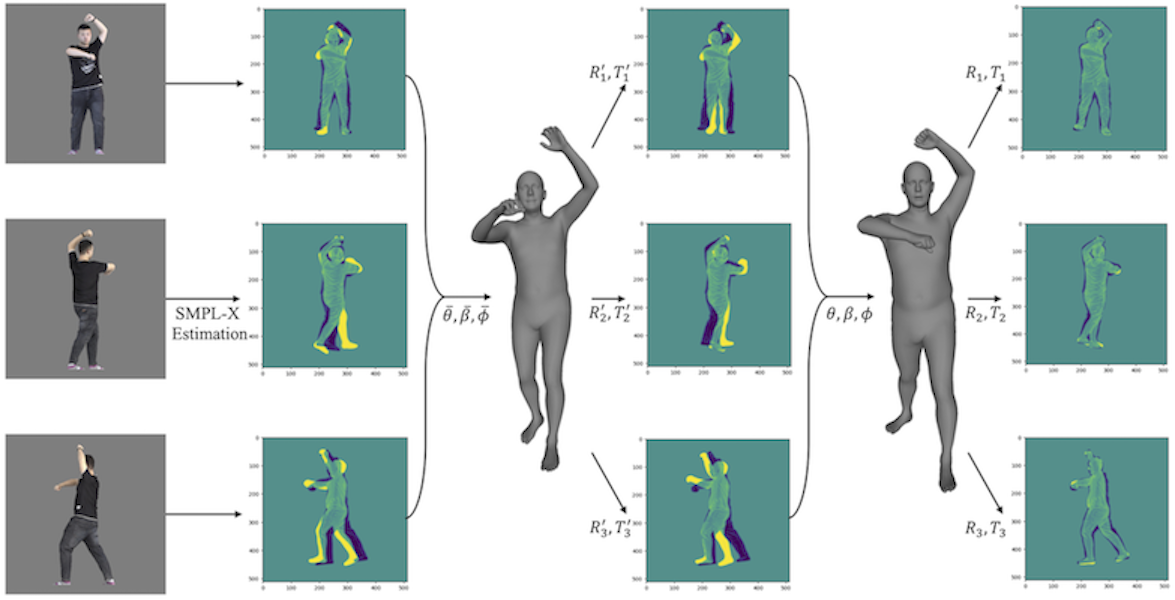}

   \caption{SMPL-X Optimization. By maximizing IoU between the body mask and the projected silhouette, and minimizing the 2D keypoint distances, we achieve optimized SMPL-X which is well-aligned with images.}
   \label{fig:sup-smplx_optimization}
\end{figure}

Given a single-view RGB image or uncalibrated multi-view RGB images as input, we first apply PIXIE~\cite{PIXIE:2021} to fit the SMPL-X model of the human subject after segmenting out the background with rembg\footnote{\url{https://github.com/danielgatis/rembg}}. 
For the multi-view case, we first fit the SMPL-X models for each input image and take the mean of the fitted shape, pose, and expression parameters $\Bar{\theta}, \Bar{\beta}, \Bar{\phi}$ as the shared model for all views, while each model keeps its own global orientation parameters $R^{'}_i, T^{'}_i$.
However, the resulting SMPL-X model is often misaligned with the image, degrading the quality of the final reconstruction. 
To remedy this problem, as described in \cref{sec:pre}, we refine the fitted SMPL-X model by maximizing the IoU between the projected SMPL-X silhouette and human body mask~\cite{cheng2022generalizable} and minimizing the 2D keypoint distances, to achieve optimized SMPL-X parameters, \ie, $\theta, \beta, \phi, R_i, T_i$ (see \cref{fig:sup-smplx_optimization}).

\paragraph{Failure cases} We initialize the SMPL-X model using all the information from multi-view images, which is generally robust. However, there may still be instances of failures when the occlusions are intense and cannot be resolved through multi-view images, particularly when the input viewpoints have short baselines.

\subsection{Learnable parameters and inference latency} 
Our model has around 18M trainable parameters and the inference time is around 20s. As a reference, PIFu has about 15M trainable parameters and requires a similar inference time.

\clearpage

\section{Further analysis}

\subsection{Effects of SeSDF module}

\begin{figure}[hbtp] \centering
    \includegraphics[width=0.97\textwidth]{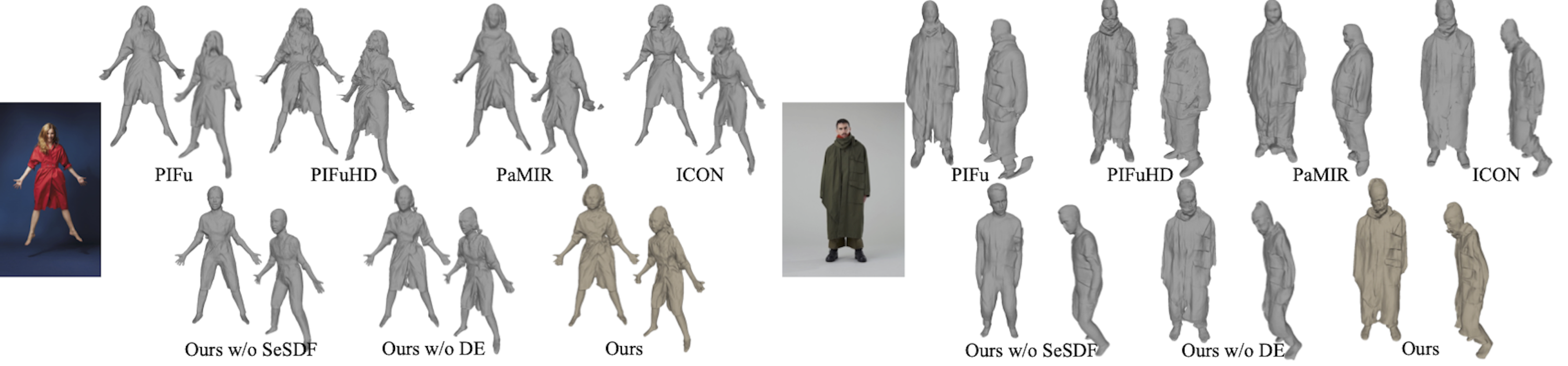}
    \caption{Comparisons on real-world images with loose clothing and easier poses. Best viewed in PDF with zoom.} \label{fig:sup-loose_clothing}
\end{figure}

Our SeSDF successfully evolves the signed distance field derived from SMPL model, by leveraging the 2D pixel-aligned feature and 3D space-aligned feature. In addition to the quantitative analysis in \cref{tab:quantitative_single} and \cref{tab:quantitative_sparse} in the main paper, we further provide qualitative comparisons on real-world images with loose clothing in~\cref{fig:sup-loose_clothing}  to analyze the visual improvement provided by our SeSDF module.

It can be observed that our SeSDF module can significantly improve the SDF computed from the `naked' SMPL-X model, and successfully reconstruct loose clothing to some extent. By comparing `Ours w/o DE' with `Ours', we can see that distance encoding can help improve the clothing details. Overall, our full model produces the best results, outperforming the other methods on real-world images with easier poses.

\paragraph{`Ours w/o SeSDF' vs ICON} `Ours w/o SeSDF' still differ from ICON in several aspects: (1) we apply Distance Encoding (DE) to SMPL-X based SDF and compute 3D point features from SMPL-X, while ICON does not contain these components; (2) ICON uses a network to predict normals from the 2D image, while our method directly takes SMPL-X vertex normals.

\clearpage

\subsection{Effects of feature design}
\begin{figure}[h]
  \centering
   \includegraphics[width=0.8\linewidth]{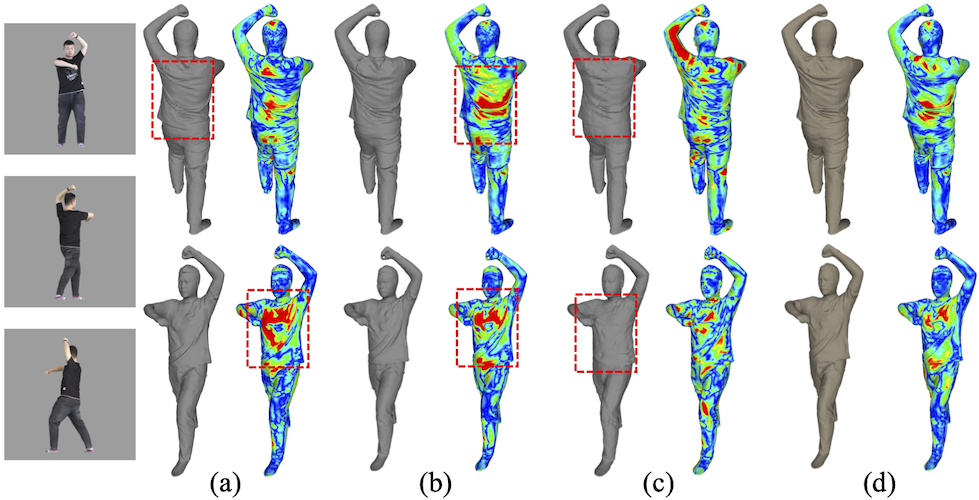}

   \caption{Further analysis of the design of our feature modules. (a) Ours w/ only the 3D feature from the first view; (b) Ours w/o 3D feature; (c) Ours w/o 2D and 3D features in \cref{ssdfunction}; (d) Ours.}
   \label{fig:sup-ablation_fd}
  \vspace{-1.2em} 
\end{figure}

SeSDF learns to predict the occupancy values for the 3D human reconstruction by \cref{ssdmodule} and \cref{ssdfunction}: 

\begin{equation}\label{ssdmodule}
    f_{sd}(F_{2D}(X), F_{3D}(X), \mathcal{D}(d'(X)), \mathbf{n}'(X))  \mapsto (d, \mathbf{n}), 
\end{equation}
\begin{equation}\label{ssdfunction}
    f_o(F_{2D}(X), F_{3D}(X), \mathcal{D}(d(X)), \mathbf{n}(X), Z(X))  \mapsto [0, 1].
\end{equation}
In our method, we use both 2D pixel-aligned ($F_{2D}(X)$) and 3D space-aligned ($F_{3D}(X)$) features for both self-evolved SDF learning (\cref{ssdmodule}) and occupancy prediction (\cref{ssdfunction}). In~\cref{fig:sup-ablation_fd},  we show results on three alternatives to further validate our choice of features. Namely, (a) instead of using 3D features from all transformed SMPL-X models by $R_i, T_i$, we only use the 3D feature from the model in the first view; (b) we exclude 3D features in \cref{ssdmodule} and \cref{ssdfunction} to demonstrate its effectiveness; (c) we eliminate the use of 2D and 3D features in \cref{ssdfunction} to validate the necessity of 2D and 3D features for occupancy prediction. As can be seen, using 3D features from the SMPL-X model under transformed views is effective, and our choice of using both 2D and 3D features for both self-evolved SDF learning and occupancy prediction appears to be optimal.

\clearpage

\subsection{Effects of feature fusion strategy}
In the multi-view reconstruction setting, the feature fusion method plays an important role. In~\cref{fig:ablation_ff}, we provide more comparison of our occlusion-aware feature fusion method with average pooling (``AVG-Pool")~\cite{saito2019pifu}, SMPL-visibility (``SMPLX-Vis")~\cite{xiu2022icon} and attention-based mechanism (``Attention")~\cite{zheng2021deepmulticap}. In addition, we also provide an alternative to our occlusion-aware feature fusion method by leveraging the surface normals rather than the depth, which can be defined as:
\begin{equation}
    N_i(X) = \tanh{(\arccos{(\frac{\mathop{v_n}\limits ^{\rightarrow}\cdot \mathop{v_d}\limits ^{\rightarrow}}{||\mathop{v_n}\limits ^{\rightarrow}||\cdot||\mathop{v_d}\limits ^{\rightarrow}||})})}.
\end{equation}
\begin{equation}
    F_{fused}(X) = \frac{\sum_{i=1}^{n} N_i(X) F_i(X)}{\sum_{i = 1}^n N_i(X)}.
\end{equation}
We call this normal-based fusion strategy ``Normal-Fusion''.
    
\begin{figure}[h]
  \centering
   \includegraphics[width=1\linewidth]{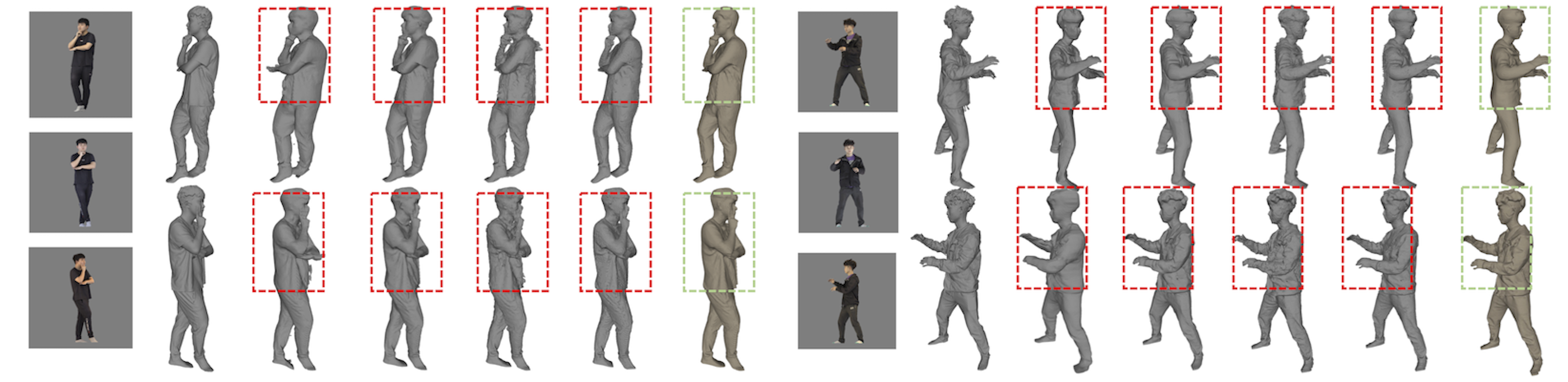}
   \includegraphics[width=1\linewidth]{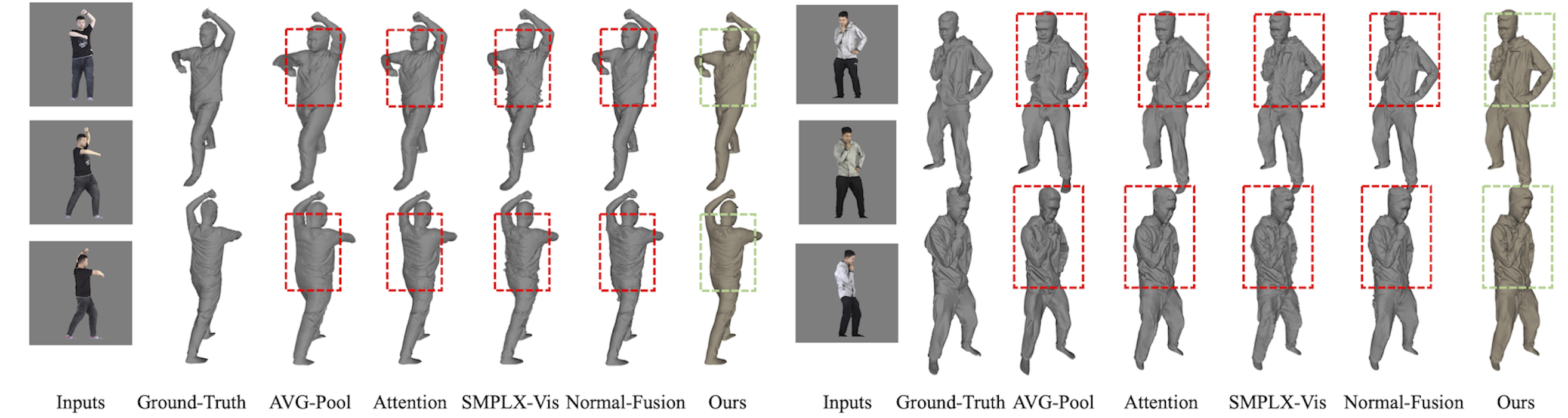}

   \caption{Comparing our occlusion-aware strategy and our normal-fusion strategy with other alternatives.}
   \label{fig:sup-ablation_ff}
  \vspace{-1.2em} 
\end{figure}

``AVG-Pool" treats different views equally, even the features from occluded views, leading to notable artifacts in the reconstruction (see Fig. 7 in the main paper). Meanwhile, it cannot handle the input images from non-evenly distributed views due to the presence of self-occlusion.

either (see \cref{fig:sup-ablation_ff}).
``SMPLX-Vis" discards the features of a 3D point $X$, if its closest SMPL-X vertex is invisible along the camera view direction. However, this strategy will negatively affect the reconstruction by introducing severe artifacts across the surface (see \cref{fig:ablation_ff} in main paper and \cref{fig:sup-ablation_ff}).
``Attention" tries to improve the features by learning the correlations across different views. Unfortunately, it also appears to suffer from the occluded features (see \cref{fig:ablation_ff} in main paper), and has difficulty handling
input images from non-evenly distributed views (see \cref{fig:sup-ablation_ff}).
``Normal-Fusion" appears to be more effective than others, except our depth-based occlusion-aware fusion method (``Ours'').
However, the occluded 3D point $X$ may also have a large $N_i(X)$ coefficient, which is undesired but often occurs in the boundary regions, leading to minor artifacts across the boundary of the 3D human reconstruction (see \cref{fig:sup-ablation_ff}). We also provide the quantitative evaluation on ``Normal-Fusion'' in the last row of \cref{tab:quantitative_sparse} in the main paper. As can be seen, our occlusion-aware fusion method based on depth consistently achieves the best performance. 

\clearpage

\subsection{Analysis of occupancy prediction}

Instead of directly applying marching cube~\cite{li2018differentiable} to the evolved signed distance field output by SeSDF module, we reconstruct the final 3D human model from the occupancy field (see \cref{fig:sup-ab_normal+ov}). Through experimental results, we find that, without the occupancy prediction, the reconstruction appears to be grainy. The occupancy prediction can greatly help smooth the reconstruction.

\subsection{Analysis of Normal}

Like ICON~\cite{xiu2022icon} which predicts per-pixel normal to aid occupancy prediction, we choose to predict normal for 3D points. The intuition behind this is that directly deriving the normal from the signed distance field will not provide additional information to improve the occupancy prediction.

\begin{figure}[hbtp] \centering
    \includegraphics[width=0.97\textwidth]{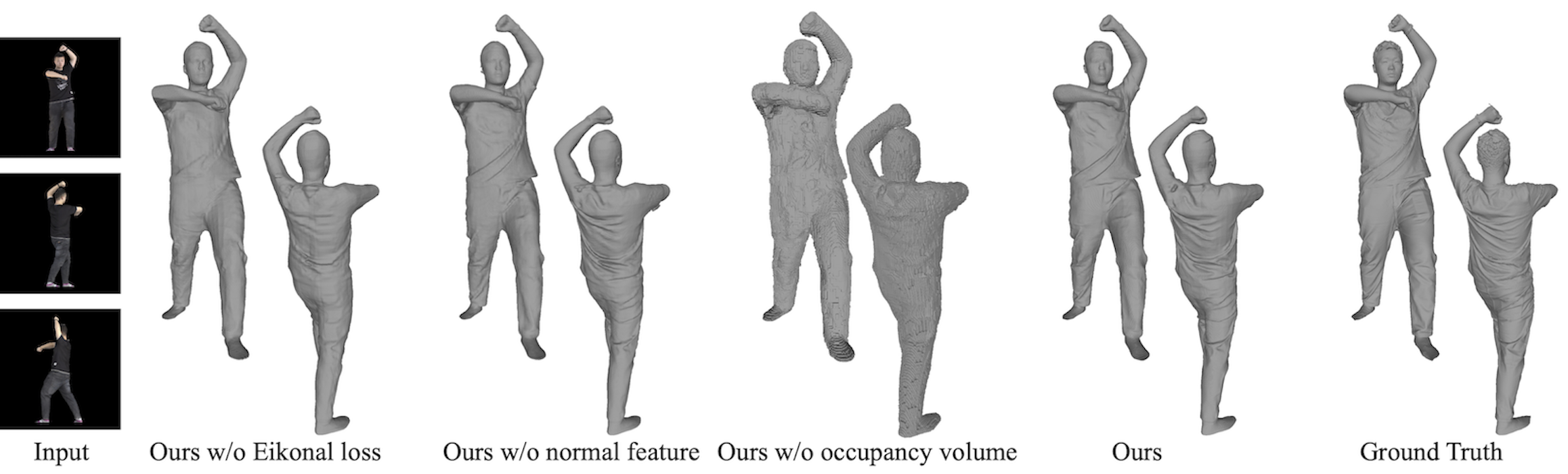}
    \caption{Ablation study for normal and occupancy volume. Best viewed in PDF with zoom.} \label{fig:sup-ab_normal+ov}
\end{figure}

In \cref{fig:sup-ab_normal+ov}, we further provide ablation analysis on the normal information. It demonstrates that the Eikonal loss is beneficial and the normal feature helps predict better occupancy value.

\clearpage

\section{More qualitative results}
\subsection{Single-view reconstrcution}
\subsubsection{Reconstruction with real-world images}
\begin{figure}[h]
  \centering
   \includegraphics[width=0.97\linewidth]{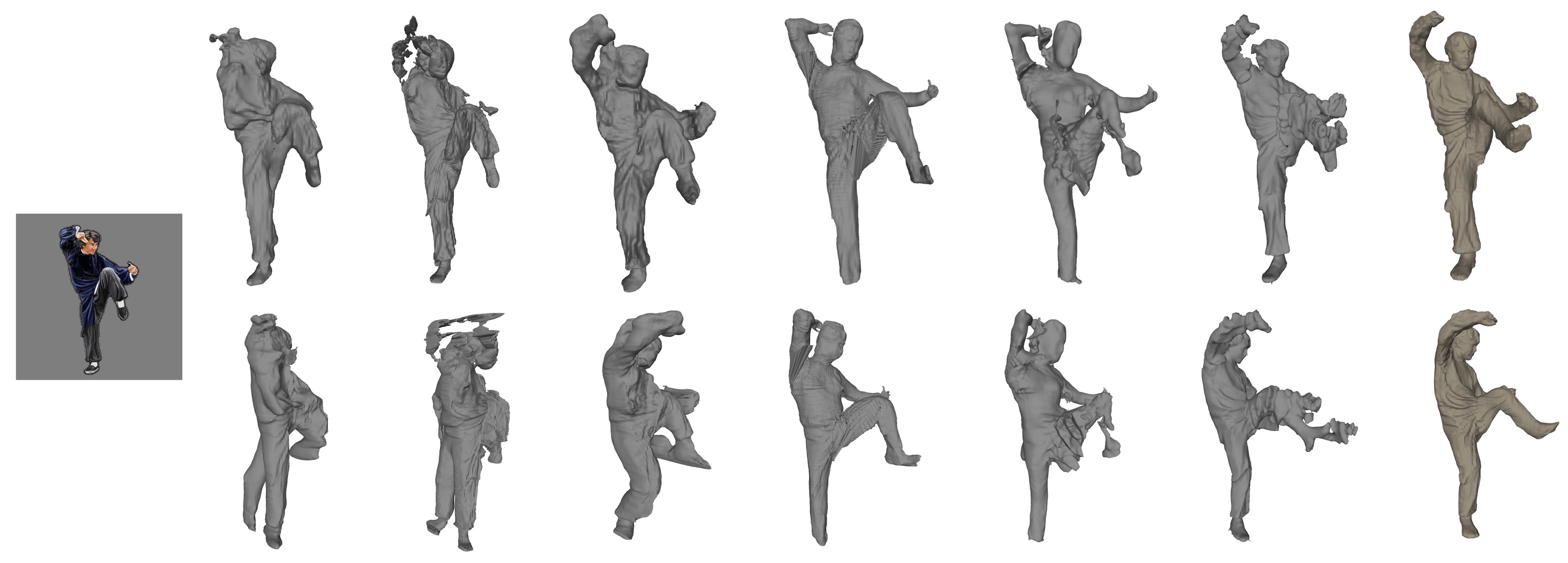}
   \includegraphics[width=0.97\linewidth]{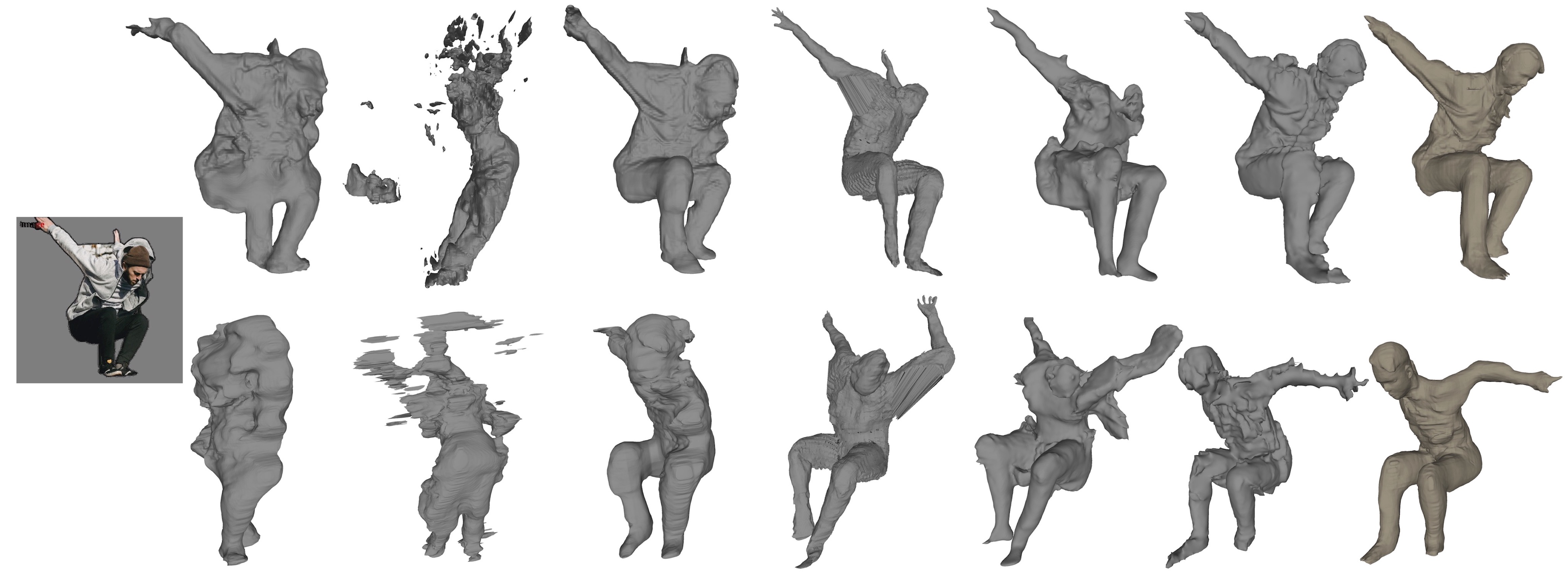}
   \includegraphics[width=0.97\linewidth]{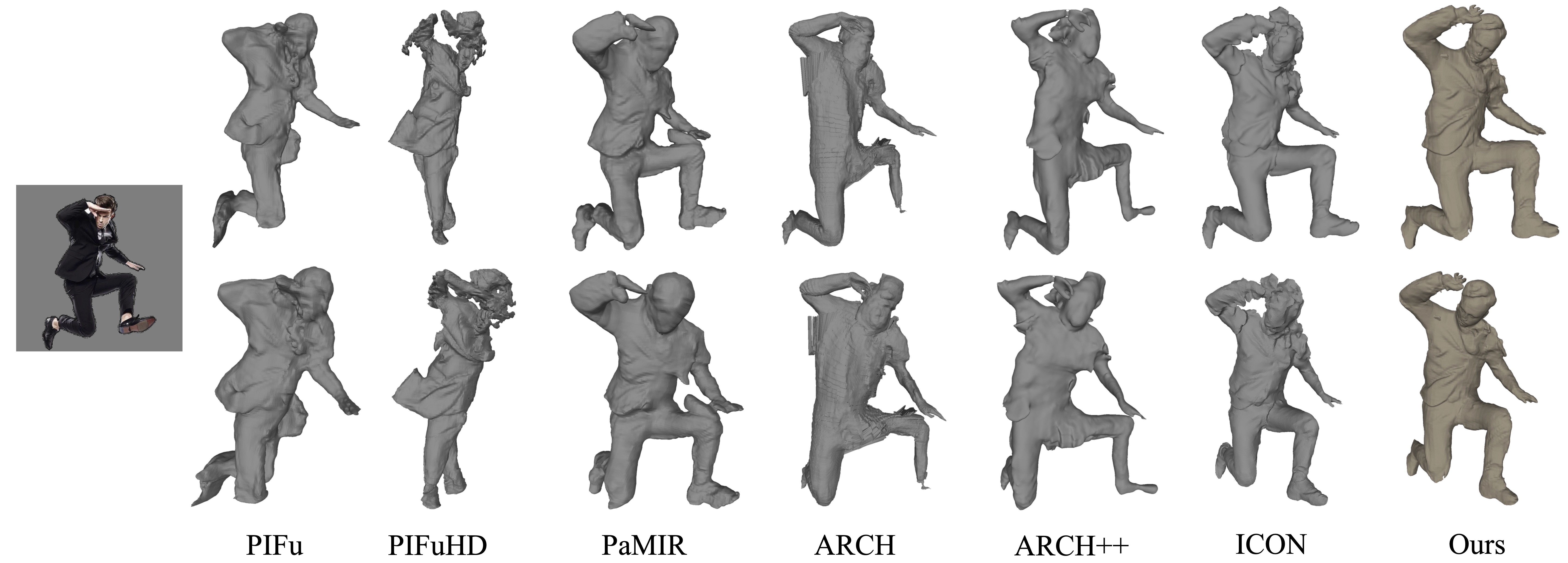}

   \caption{Qualitative comparison with SOTA methods on real-world images.}
   \label{fig:sup-real-1}
  \vspace{-1.2em} 
\end{figure}

\begin{figure}[h]
  \centering
   \includegraphics[width=1\linewidth]{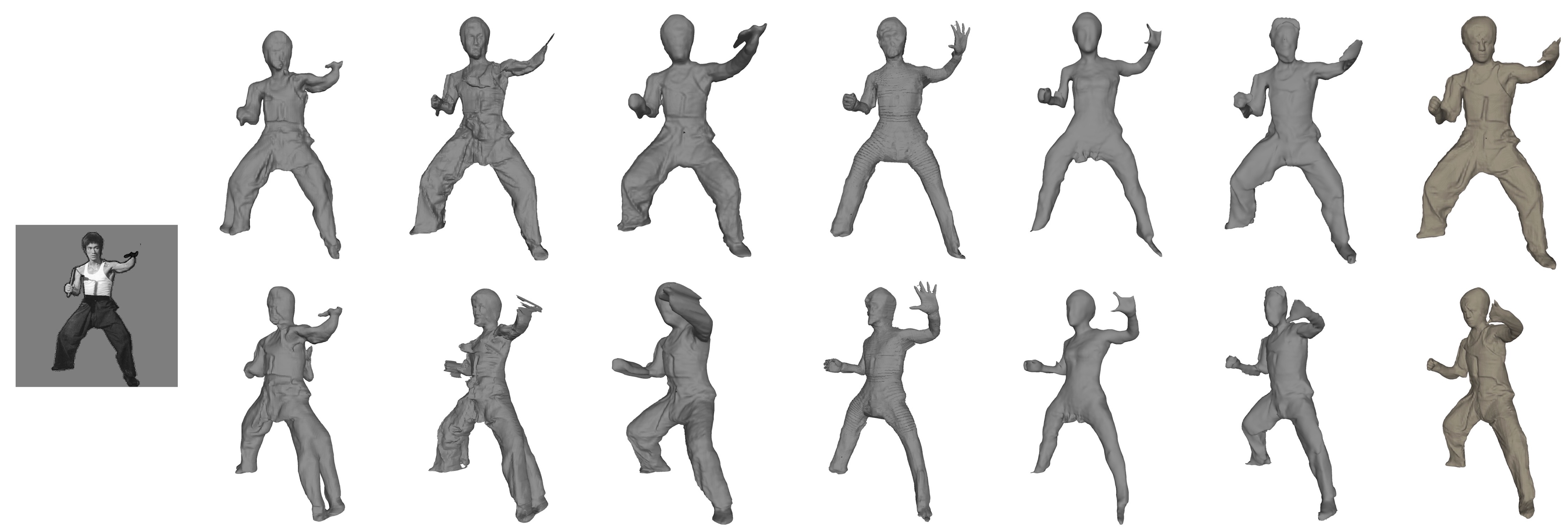}
   \includegraphics[width=1\linewidth]{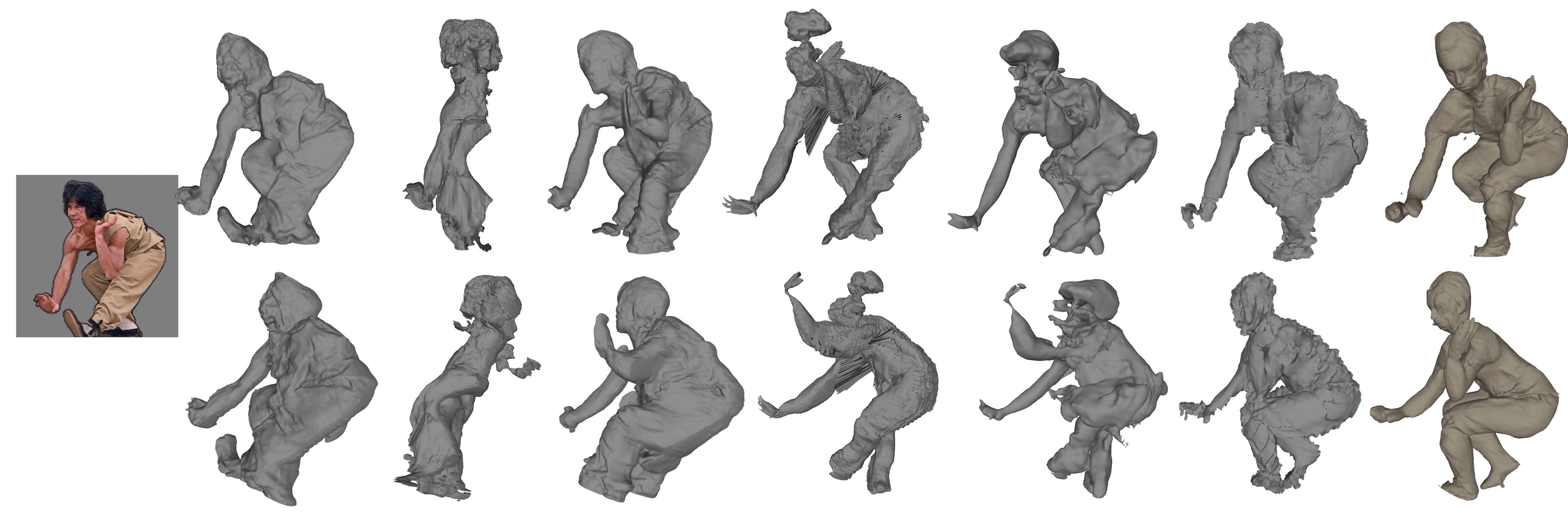}
   \includegraphics[width=1\linewidth]{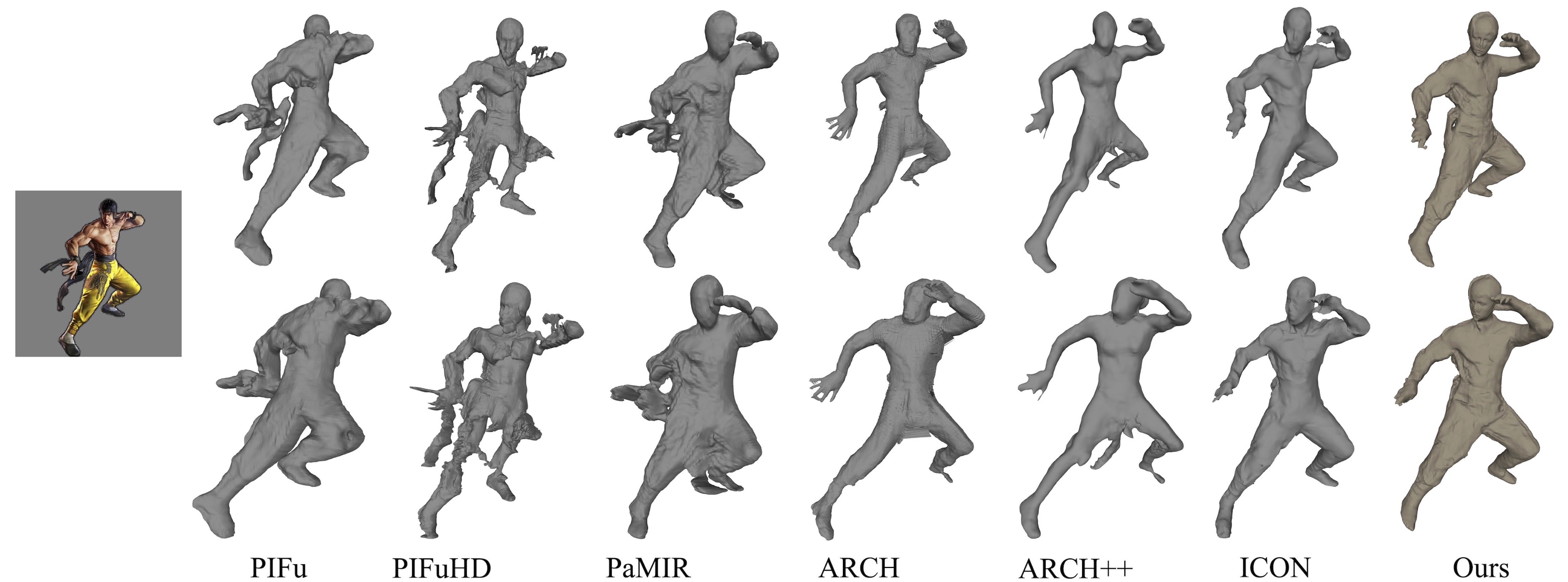}

   \caption{Qualitative comparison with SOTA methods on real-world images.}
   \label{fig:sup-real-2}
  \vspace{-1.2em} 
\end{figure}

\begin{figure}[h]
  \centering
   \includegraphics[width=1\linewidth]{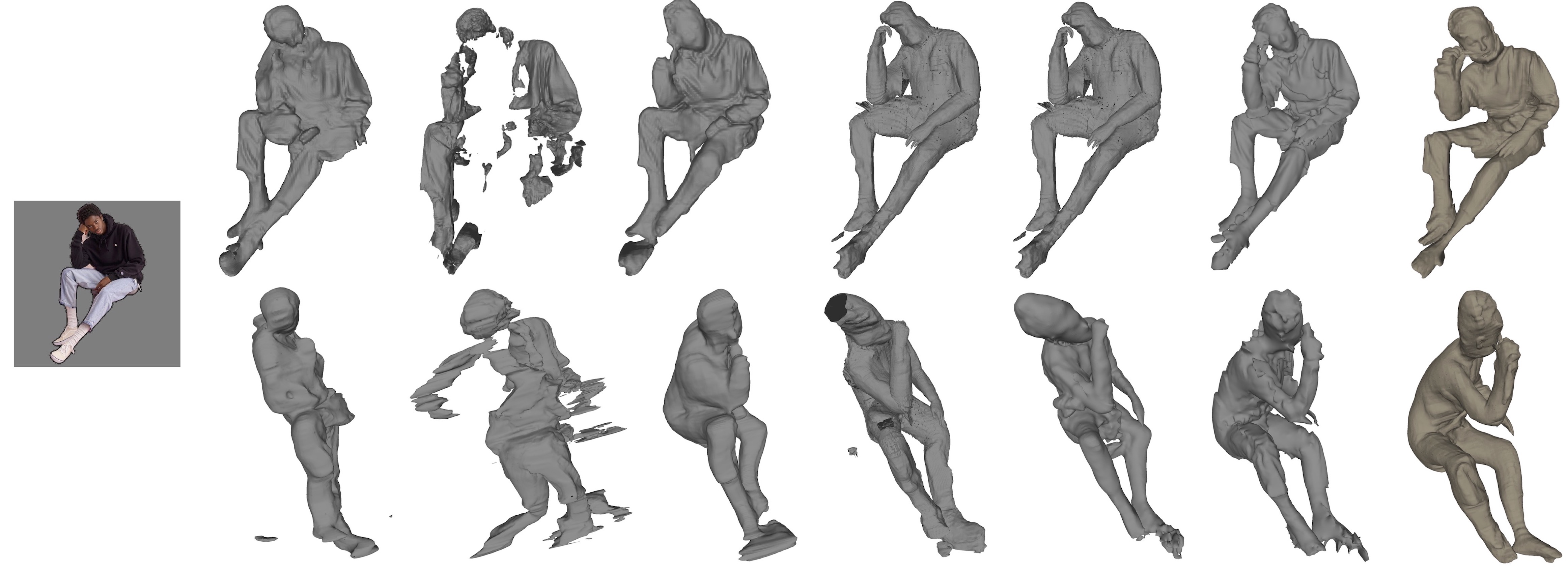}
   \includegraphics[width=1\linewidth]{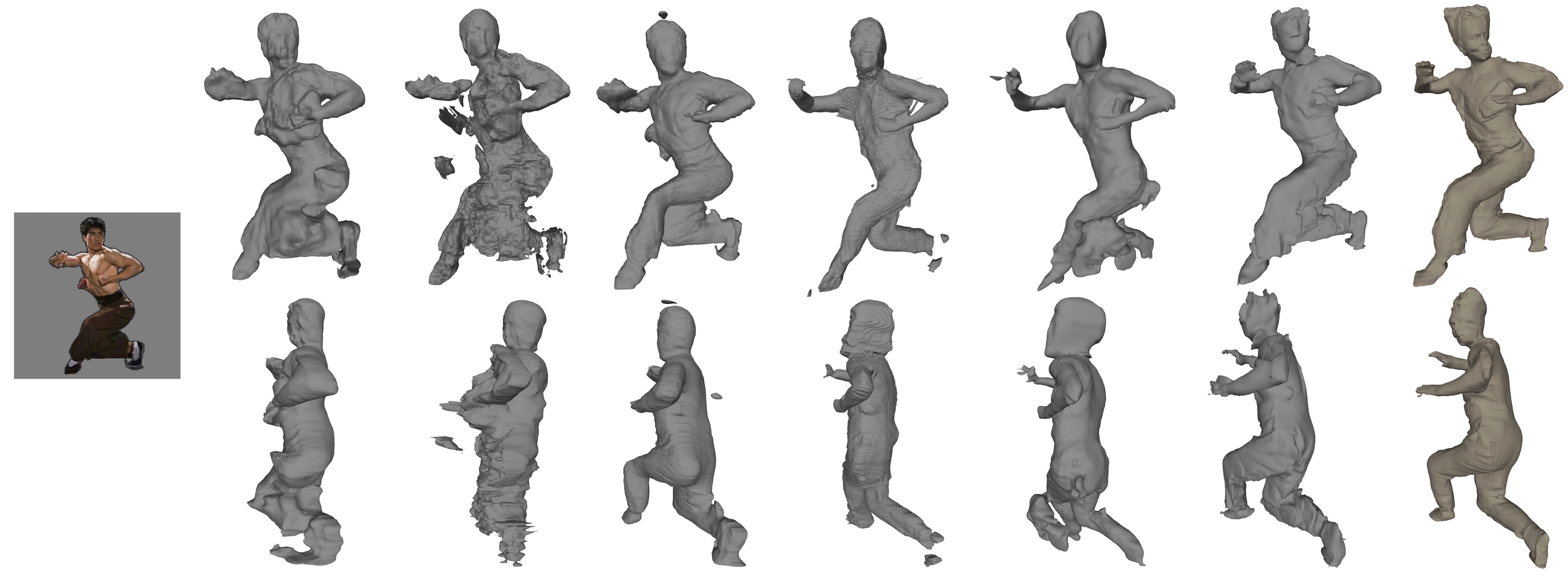}
   \includegraphics[width=1\linewidth]{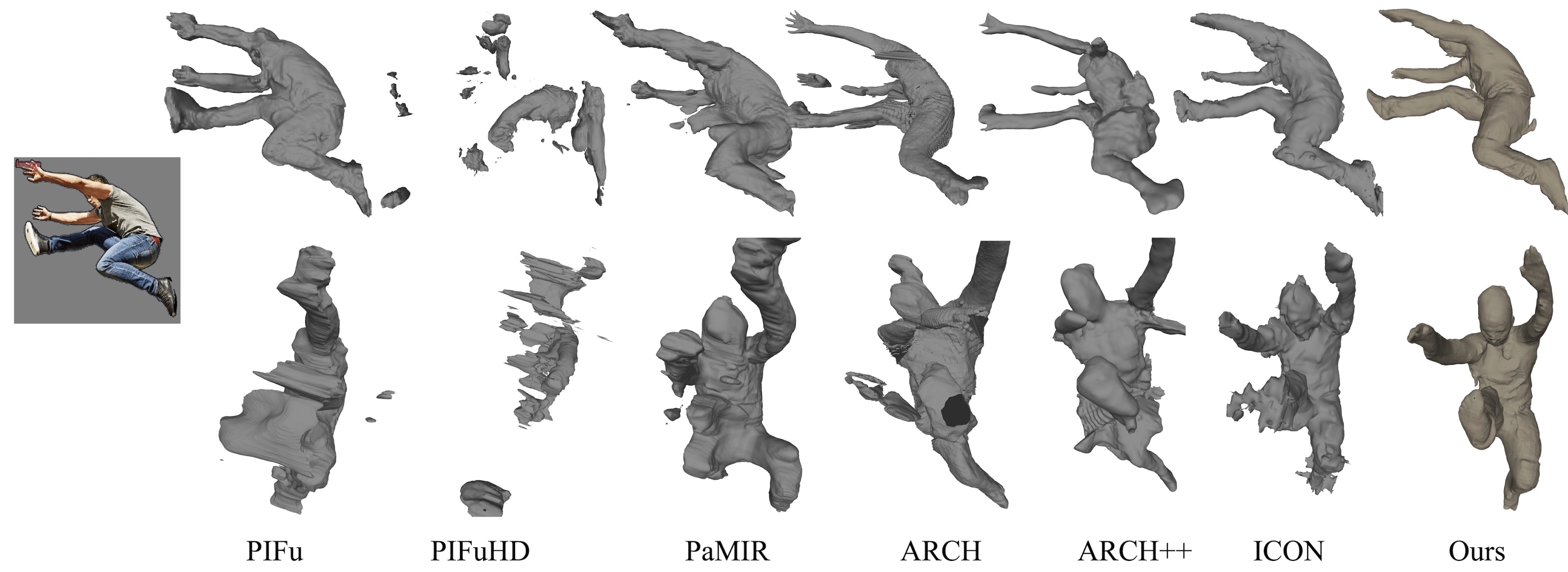}

   \caption{Qualitative comparison with SOTA methods on real-world images.}
   \label{fig:sup-real-3}
  \vspace{-1.2em} 
\end{figure}

\begin{figure}[h]
  \centering
  \includegraphics[width=1\linewidth]{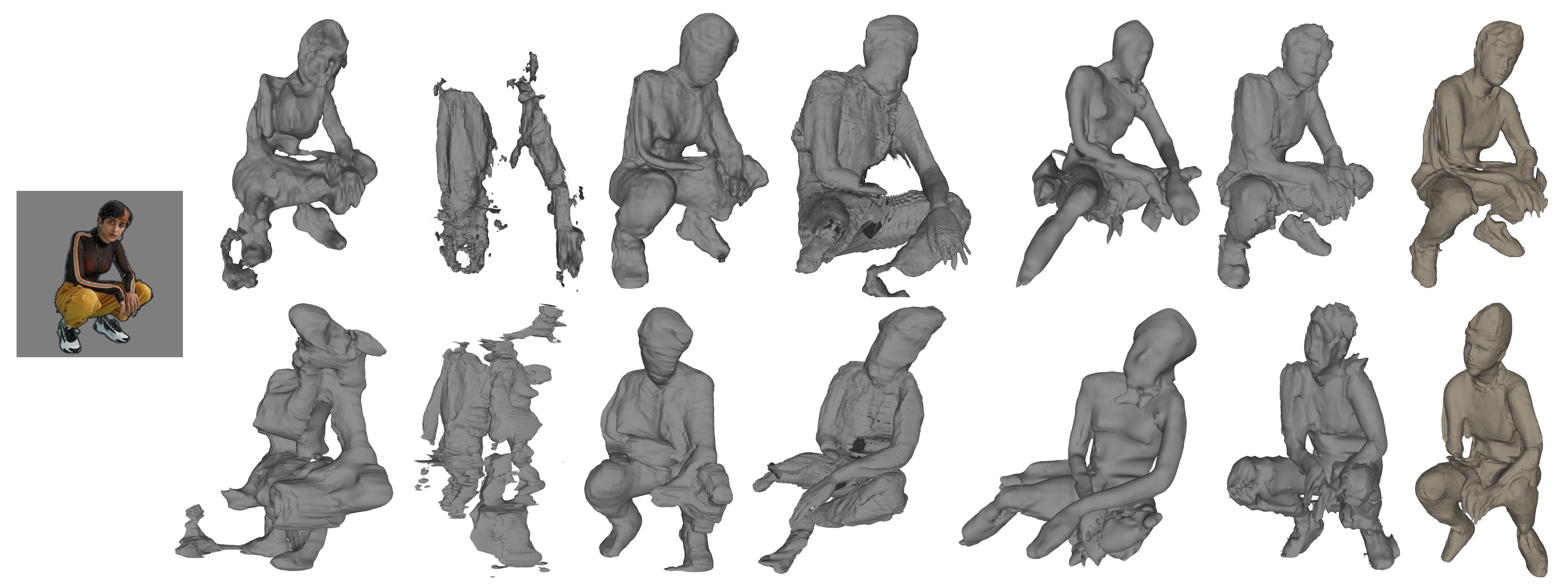}
   \includegraphics[width=1\linewidth]{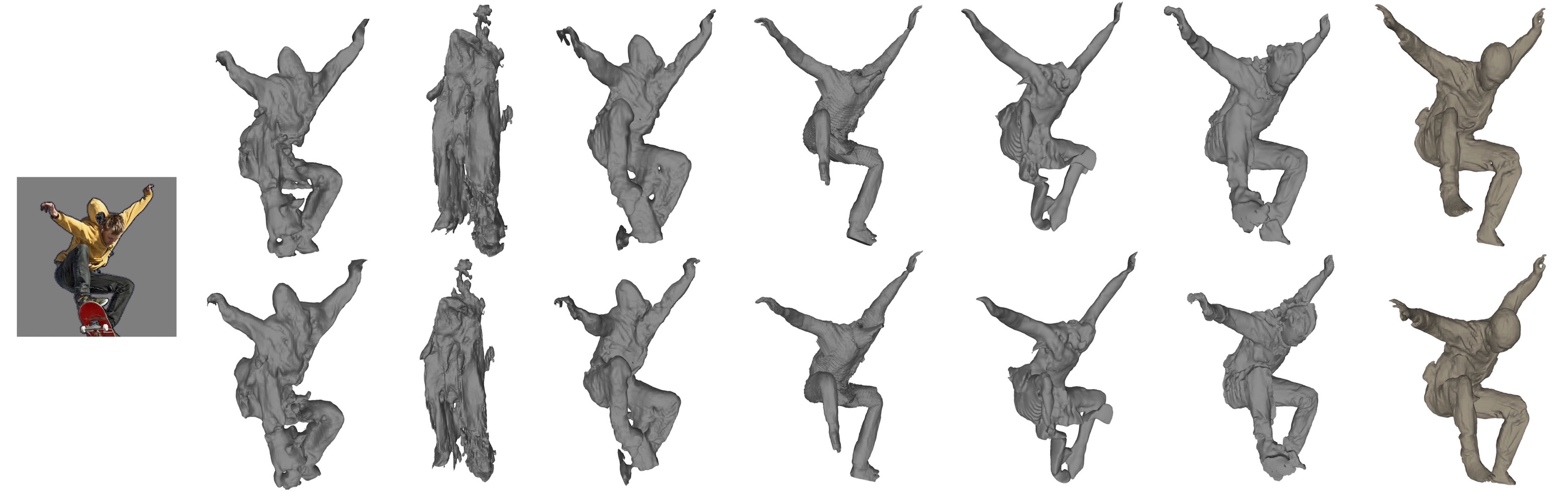}
   \includegraphics[width=1\linewidth]{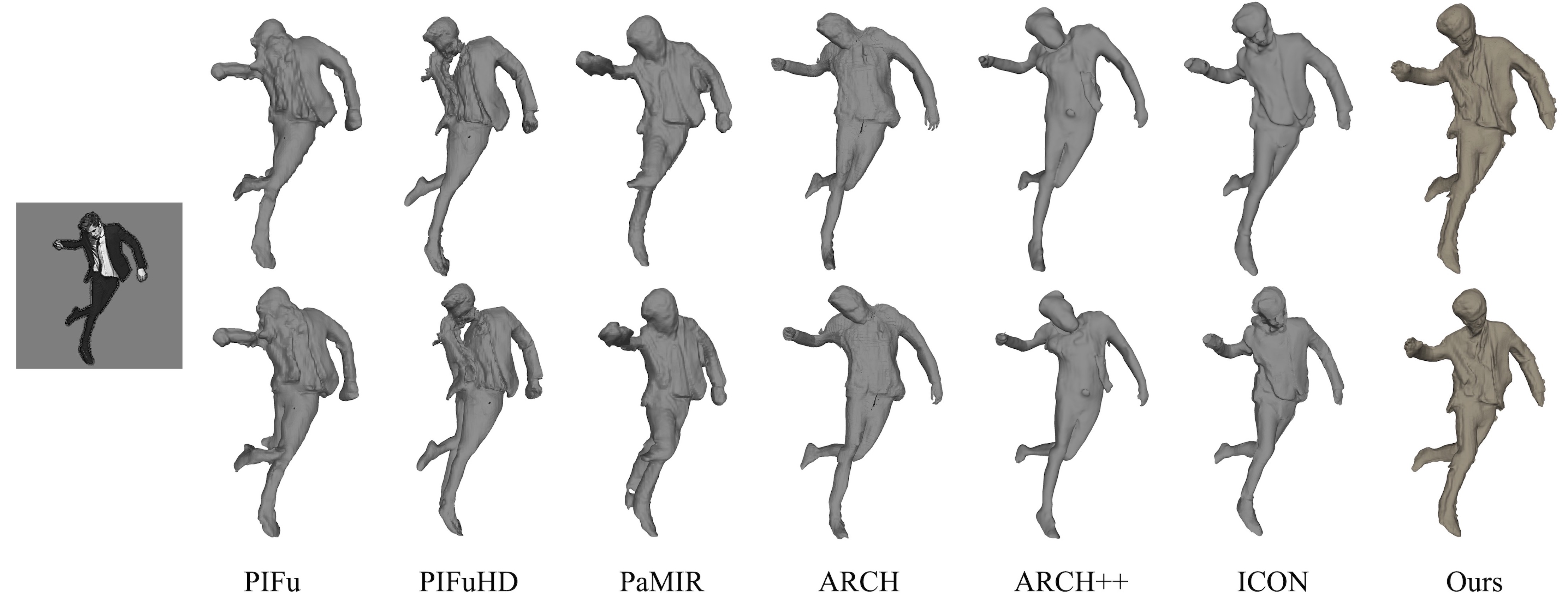}

   \caption{Qualitative comparison with SOTA methods on real-world images.}
   \label{fig:sup-real-4}
  \vspace{-1.2em} 
\end{figure}

\cref{fig:sup-real-1} - \cref{fig:sup-real-4} show more qualitative comparisons between our SeSDF and other SOTA methods on real-world images.
As can be observed, our method can robustly handle complicated poses, and faithfully reconstruct fine details.
\clearpage

\clearpage

\clearpage

\clearpage

\subsubsection{Reconstruction with THUman2.0 images}
\begin{figure}[h]
  \centering
   \includegraphics[width=1\linewidth]{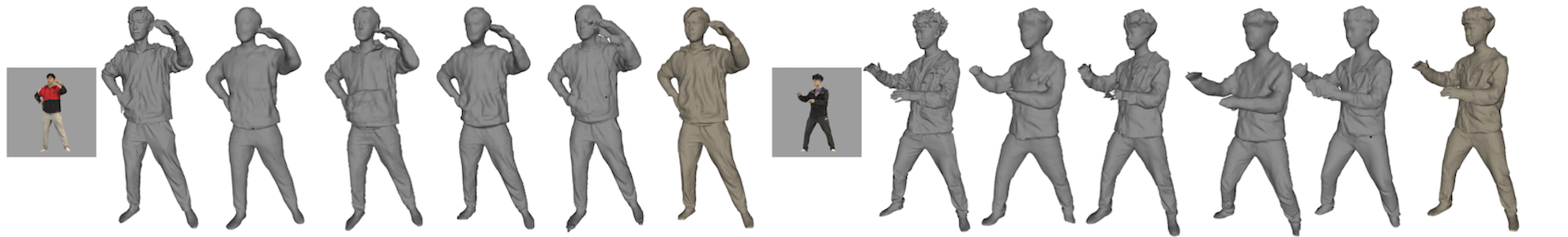}
   \includegraphics[width=1\linewidth]{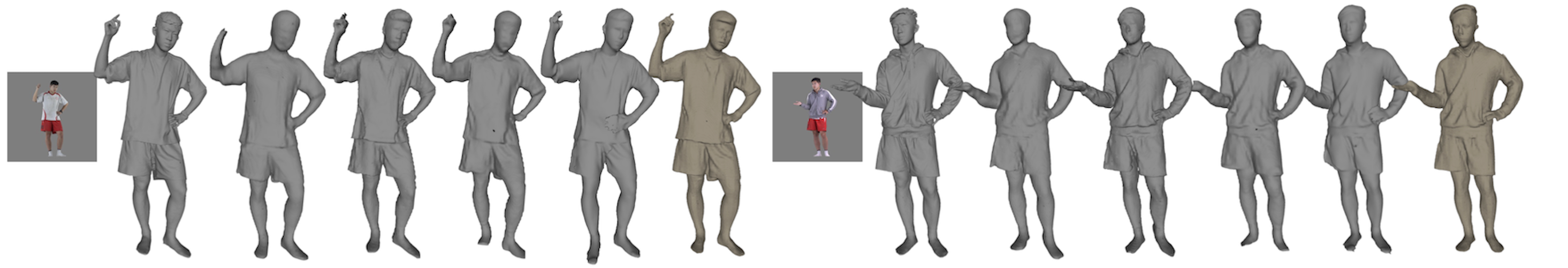}
   \includegraphics[width=1\linewidth]{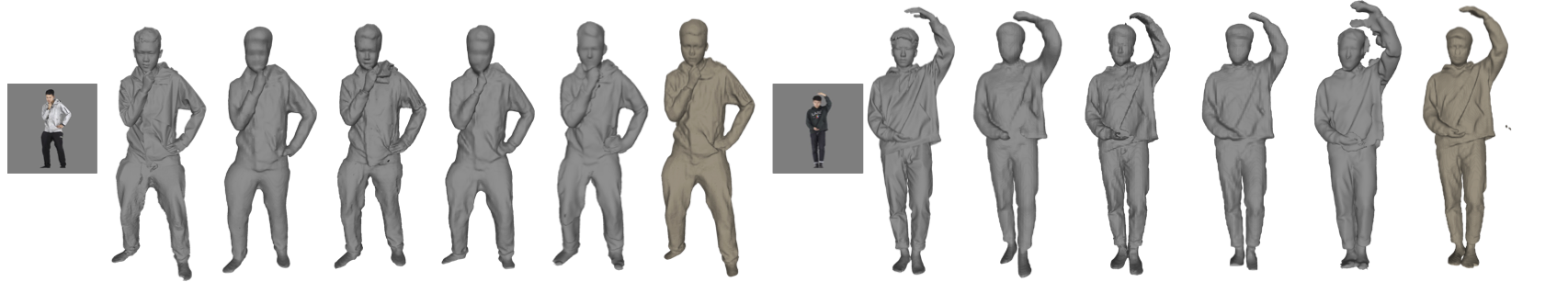}
   \includegraphics[width=1\linewidth]{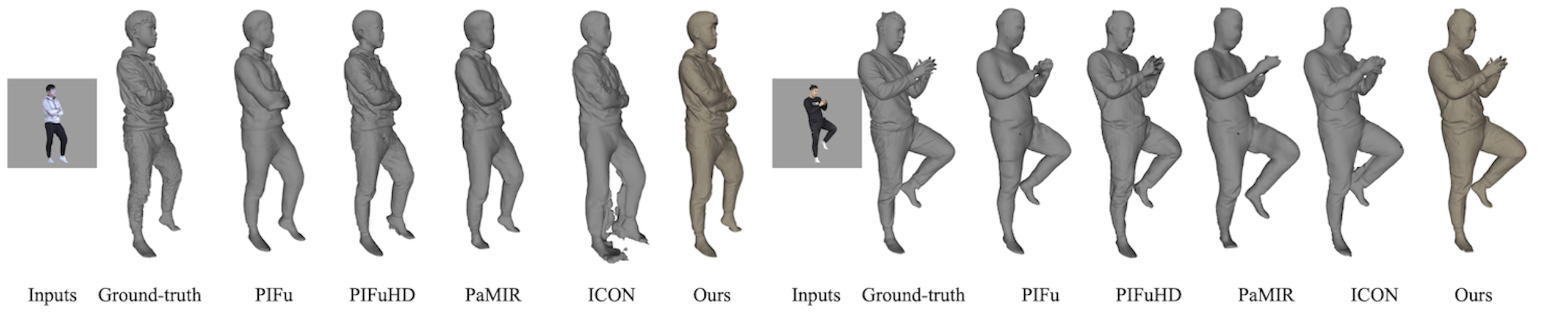}

   \caption{Qualitative comparison with SOTA methods on THUman2.0 test data with controlled poses.}
   \label{fig:sup-thu2}
  \vspace{-1.2em} 
\end{figure}

\cref{fig:sup-thu2} provides more comparisons on THUman2.0 test data. It demonstrates that our method can consistently reconstruct better clothing topology than existing SOTA methods under different scenarios.
\clearpage

\subsection{Improvement from single-view to multi-view reconstruction}

\begin{figure}[h]
  \centering
  \includegraphics[width=1\linewidth]{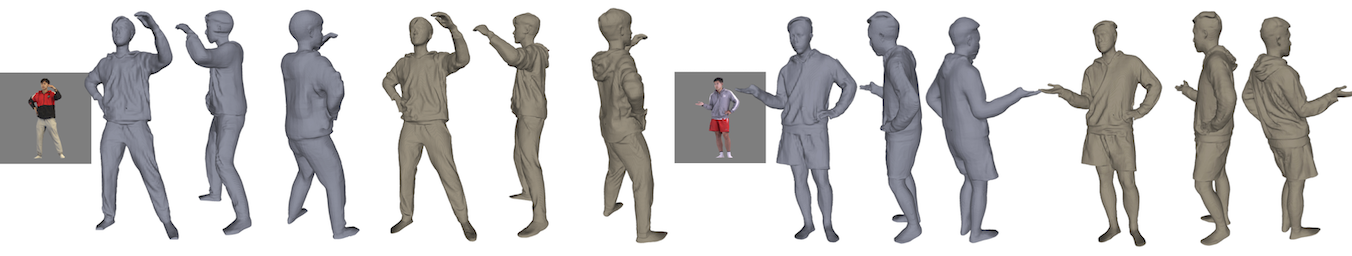}
   \includegraphics[width=1\linewidth]{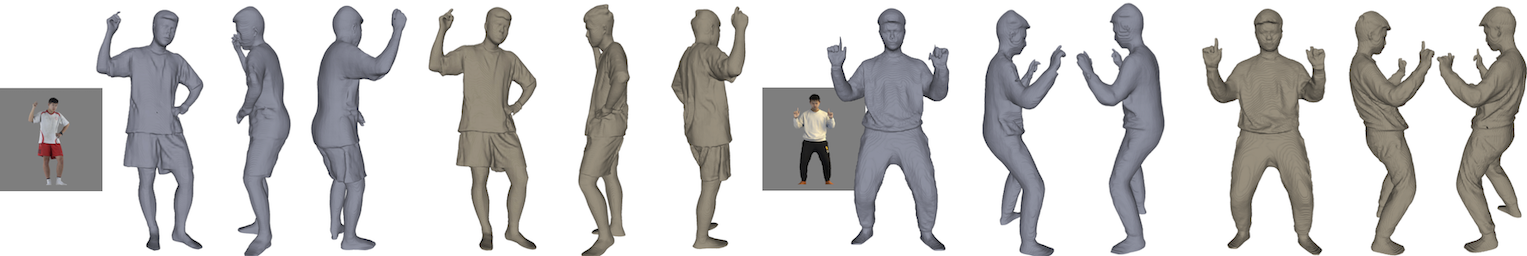}

   \caption{\textcolor{blue}{Single-view reconstruction} vs \textcolor{orange}{multi-view reconstruction}.}
   \label{fig:sup-single_to_multi}
  \vspace{-1.2em} 
\end{figure}

In~\cref{fig:sup-single_to_multi}, we show results of the same subject using a single image or uncalibrated multi-view images (we use three images) as the input for SeSDF. As can be seen, more details, especially those that are absent from the single-view input, can be reconstructed faithfully by our SeSDF framework.

\clearpage
\subsection{Multi-view reconstruction}
\subsubsection{Reconstruction on THUman2.0}
\begin{figure}[h]
  \centering
   \includegraphics[width=1\linewidth]{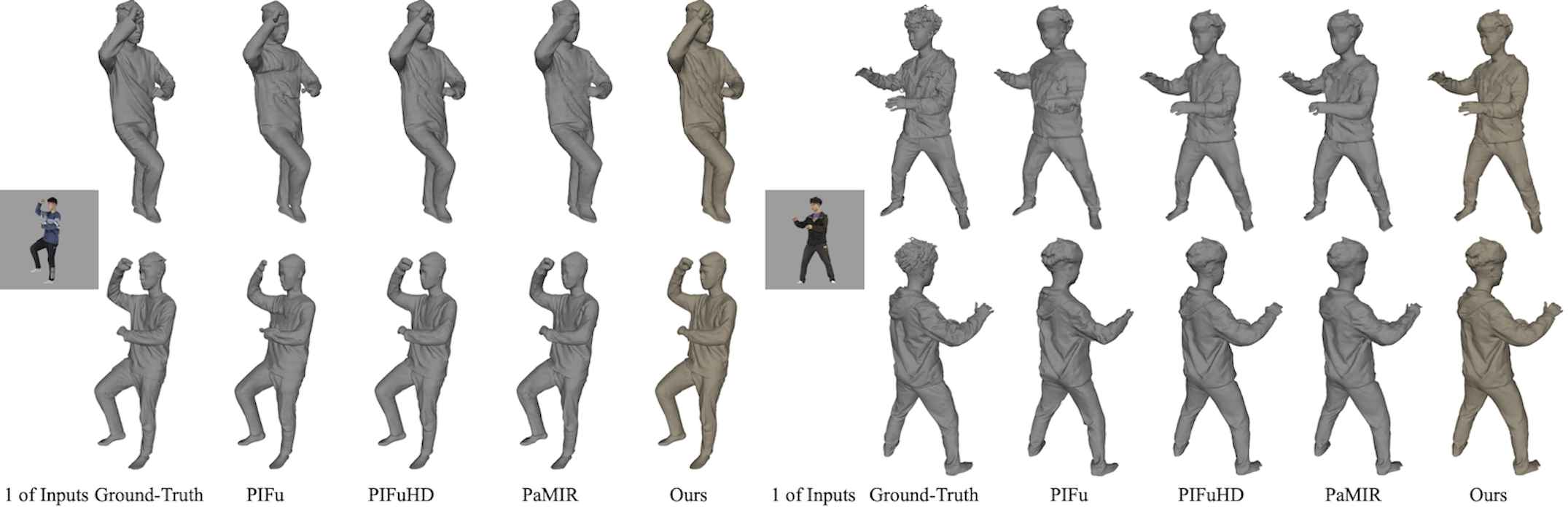}

   \caption{Qualitative comparison with SOTA methods on THUman2.0 test data for multi-view 3D human reconstruction.}
   \label{fig:sup-multi}
  \vspace{-2em} 
\end{figure}

We provide more multi-view reconstruction results in \cref{fig:sup-multi} to demonstrate the performance of our method.

\subsubsection{Different number of views}
Our SeSDF framework can be trained and tested on an arbitrary number of views and the views during the training and testing are not necessarily to be equal. In \cref{fig:sup-nviews-1} - \cref{fig:sup-nviews-2}, we train our network with 3-views and test with different numbers of views:
\begin{figure}[h]
  \centering
   \includegraphics[width=1\linewidth]{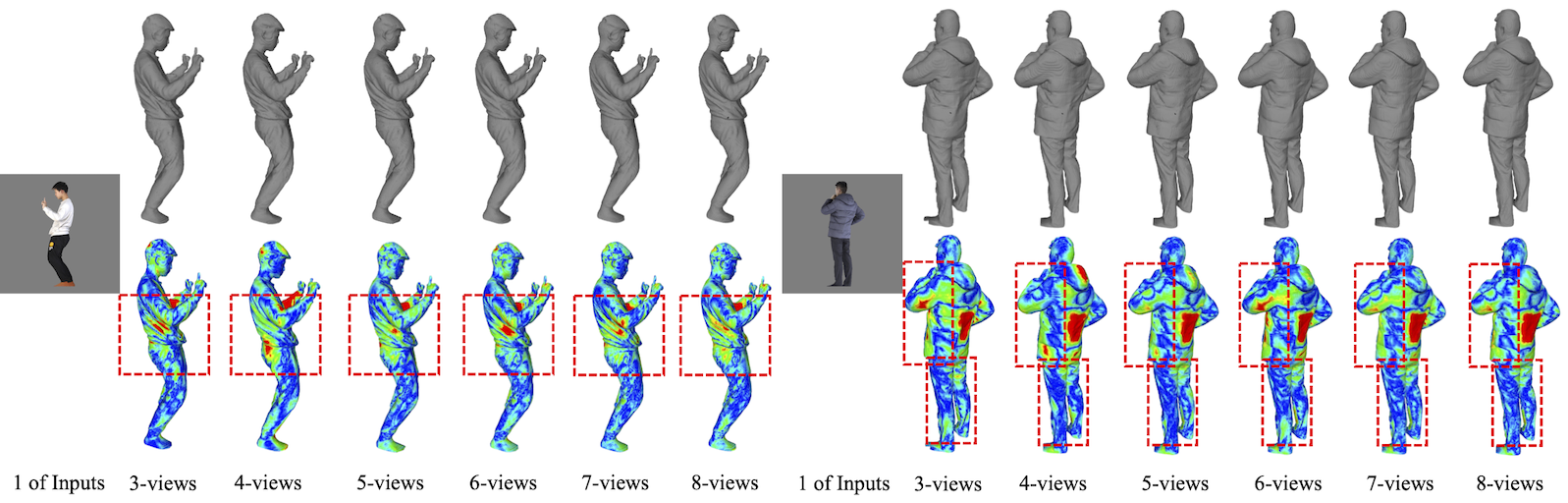}

   \caption{Reconstructions with different numbers of input views.}
   \label{fig:sup-nviews-1}
  \vspace{-1.2em} 
\end{figure}
\clearpage

\begin{figure}[h]
  \centering
  \vspace{-2em}
  \includegraphics[width=0.95\linewidth]{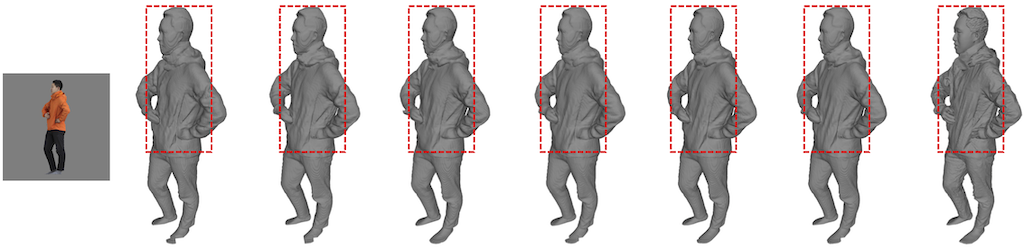}
  \includegraphics[width=0.95\linewidth]{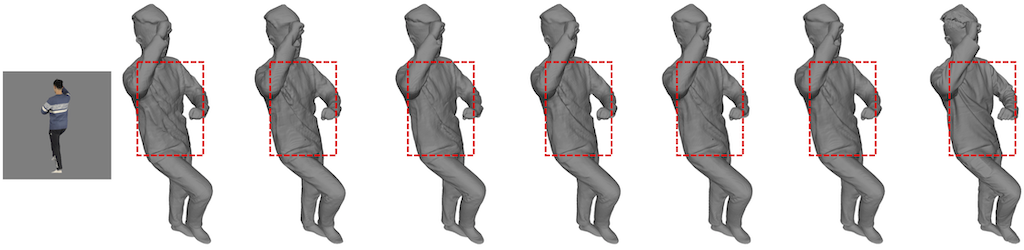}
  \includegraphics[width=0.95\linewidth]{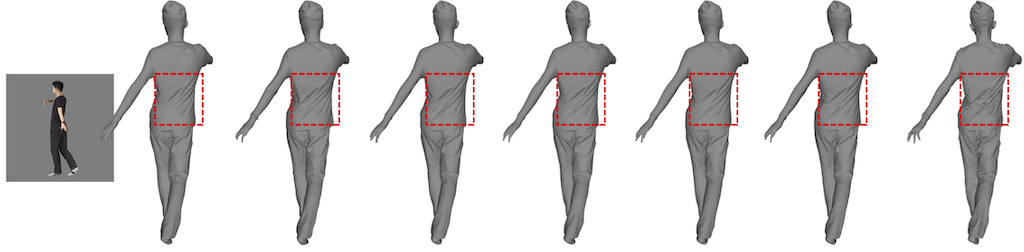}
  \includegraphics[width=0.95\linewidth]{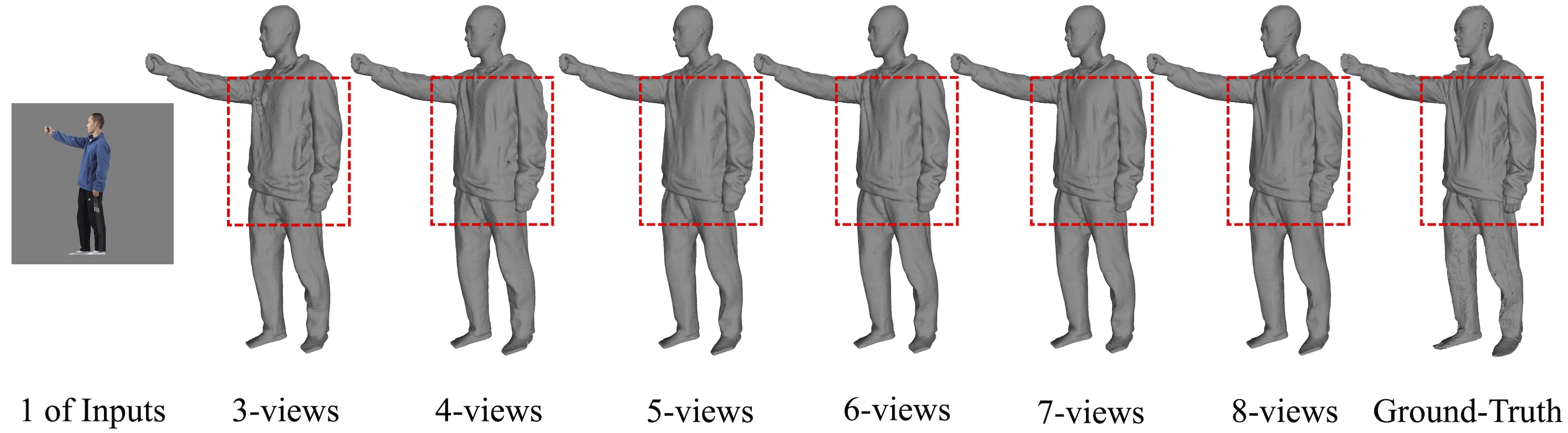}

  \caption{Reconstructions with different numbers of input views.}
  \label{fig:sup-nviews-2}
  \vspace{-2em} 
\end{figure}

\end{sloppypar}
\end{document}